%% file: main.tex
\documentclass[10pt]{article} % For LaTeX2e
\usepackage[preprint]{tmlr}

\input{math_commands.tex}

\usepackage{url}

\usepackage[hidelinks]{hyperref}
\usepackage{algorithm}
\usepackage[noend]{algpseudocode}
\algrenewcommand\algorithmicindent{0.7em}
\usepackage{xfrac}
\usepackage{subcaption}
\usepackage{wrapfig}
\allowdisplaybreaks
\setcitestyle{citesep={,},aysep={\,}}
\usepackage{makecell}
\usepackage{booktabs}
\usepackage{standalone}
\usepackage{tikz}
\usetikzlibrary{arrows.meta,bending,calc,backgrounds,fit,positioning}

\title{Streaming Recurrent Memory}

\author{\name Mohamed Elsayed \email mohamedelsayed@ualberta.ca \\
\addr Department of Computing Science\\
University of Alberta\\
\AND Minh Pham \email minh.pham@ualberta.ca \\
\addr Department of Computing Science\\
University of Alberta\\
\AND A.\ Rupam Mahmood \email armahmood@ualberta.ca \\
\addr Department of Computing Science\\
University of Alberta\\
CIFAR Canada AI Chair, Amii
}

\def\month{MM}  % Insert correct month for camera-ready version
\def\year{YYYY} % Insert correct year for camera-ready version
\def\openreview{\url{https://openreview.net/forum?id=XXXX}} % Insert correct link to OpenReview for camera-ready version

\begin{document}

\maketitle

\begin{abstract}
Natural intelligence learns from a continuous stream of sensorimotor experience: data can arrive at high bandwidth, and behavior adapts to feedback as soon as it arrives, even when it is delayed.
In contrast, most modern deep learning systems rely on stored datasets, replay buffers, and retained contexts through which credit is later assigned.
At high data rates, storing and revisiting past experience can be impractical under tight memory and computational constraints.
This paper studies recurrent memory under a streaming learning constraint: the learner learns from experience as soon as it arrives without storing raw experience.
We ask whether a streaming learner can maintain a state that remains useful for future prediction and control from delayed feedback.
We introduce Streaming Recurrent Memory (SRM), a recurrent memory designed for the streaming setting which preserves information that may matter later while remaining trainable with one-step backpropagation through time (BPTT). 
SRM provides a streaming-compatible recurrence that is designed to recall past information by content or time, track elapsed steps, and compose transformations whose outcome depends on the order in which inputs are observed. 
Across controlled memory tasks and streaming applications, SRM learns effectively with one-step BPTT, outperforming recurrent baselines in associative recall and online language modeling while achieving near-perfect success in associative T-Maze with delayed rewards.
These results suggest that fixing how inputs are stored in the recurrent state, while learning how that state is read out, enables effective streaming learning without long-range backpropagation.
\end{abstract}

\section{Introduction}

\begin{figure}[t]
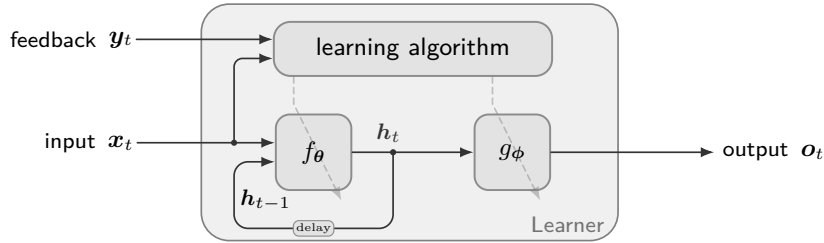

    \centering
    \includestandalone[width=0.7\linewidth]{figures/streaming-learner-tikz}
    \caption{The streaming learner. The input and feedback arrive sequentially, and the learner updates based on each sample as it arrives without storing. Feedback $\vy_t$ may be instructive, such as labels or targets in supervised learning, or evaluative, such as rewards in reinforcement learning, and often arrives delayed. The output $\vo_t$ can be a prediction or an action. The learning algorithm updates the parameterized recurrent dynamics function $f_{\boldsymbol\theta}$ and the parameterized output function $g_{\boldsymbol\phi}$.}
    \label{fig:streaming-learner}
    \vspace{-0.5cm}
\end{figure}

Many natural data sources are streams: text, audio, video, and sensorimotor experience unfold continuously, with temporal structure spanning arbitrarily long intervals. Modern sequence learners, particularly Transformers \citep{vaswani2017attention}, are instead typically trained on shuffled collections of stored segments. This strategy exploits parallel hardware and underlies state-of-the-art results across language and vision \citep{brown2020language, hoffmann2022training, touvron2023llama1}, but does not preserve the original data stream. A natural alternative is to learn directly from the stream itself: experience arrives once, in order, and the learner must learn without relying on stored history. This setting is not only close to how natural agents learn from continuous, non-repeatable experience, but also matches the constraints faced by many deployed systems, including real-time control and robotics \citep{vasan2024deep, verma2023autonomous}, never-ending video prediction \citep{carreira2024learning, qian2024streaming, han2025learning}, continual learning \citep{yoo2024lifelong}, and adaptive interaction \citep{dai2025multi}. We study this setting using recurrent learners, which process one input at a time while maintaining a state that carries information from past inputs into future computations and is used to produce predictions or actions. Figure~\ref{fig:streaming-learner} illustrates the streaming setting studied in this paper: inputs and feedback arrive sequentially, where the learner updates its recurrent state, which carries information from the past, and learns using the provided feedback and its current input.

Learning directly from a stream requires more than compressing a long history into a non-growing state \citep{hochreiter1997long}. At each moment, the learner does not yet know which parts of the past \textit{will} matter: it may later need to recall an observation, retrieve a value associated with a key, estimate elapsed time, reset a learned context, track a state progression, or assign delayed feedback to the state that produced it \citep{sutton1988learning, schmidhuber1992learning, graves2014neural, weston2014memory, sukhbaatar2015end, schlag2021linear, merrill2024illusion}. For example, a robot navigating a room may need to remember where it saw an object, bind that object to a location, track its pose, reset context when entering a new room, and credit earlier decisions for later success or failure \citep{parisotto2017neural}. Thus, the recurrent state must both preserve useful structure from the past and make that structure usable by later learning signals. Both requirements must be satisfied for effective learning; a memory whose update is easy to learn may not preserve the information needed later, while a memory that preserves information may still be useless if delayed feedback cannot shape how it is used.  The streaming setting, therefore, places a stronger constraint on recurrent architectures as their states must not only be expressive enough to store the right structure, but also be learnable from feedback observed in the stream, even when delayed.

Modern recurrent and state-space networks are a natural starting point because they maintain a non-growing state that can be updated step by step (e.g., \citealt{orvieto2023resurrecting, gu2024mamba, de2024griffin}). However, streaming learning restricts how a recurrent state update can be learned. In offline sequence learning, the learner can learn how information is written into the recurrent state through gradients propagated across a retained window using backpropagation through time (BPTT). If information written at one time is useful later, the later feedback can shape the earlier writing mechanism to incorporate that information. Streaming learning rules out this retained window. Real-Time Recurrent Learning (RTRL; \citealt{williams1989rtrl}) can recover these gradients online, but it must store and update the sensitivity of the state to every parameter, which is tractable only for restricted recurrences, and in practice, it is only approximate when recurrent layers are stacked \citep{zucchet2023online}, defeating its goal of computing exact gradients. In this work, we study streaming learning with BPTT truncated to one step instead. Under one-step BPTT, the current feedback can train how to use the current state, but it cannot directly correct the specific past update that produced that state. As a result, architectures whose useful memory depends on a learned writing mechanism face a credit-assignment bottleneck. This setting, therefore, calls for a different design principle: the state should include as much information as possible, while learning should act mainly on how that state is read.

We introduce \textit{Streaming Recurrent Memory} (SRM), a recurrent unit built around this principle. SRM constructs memory with a fixed writing mechanism that stores each input in a structured content--time address space instead of learning what to write from later feedback, a gradient path absent in one-step BPTT. 
The learner is focused on how to read from the memory, where gradients are available immediately: the state preserves information as the stream arrives, while the learner learns how to address and use that state for prediction or control. We evaluate SRM through capability probes, credit-assignment diagnostics, controlled memory tasks, and streaming applications, including next-byte language modeling and streaming deep reinforcement learning. Overall, this paper makes three contributions: we identify why streaming learning with one-step BPTT favors a memory whose writing mechanism does not depend on delayed credit, we introduce a content--time associative memory with a separate tracking channel for order-sensitive computation, and we show that this design learns memory-dependent behavior under one-step BPTT across controlled tasks, reinforcement learning, and language modeling.

%%%%%%%%%%%%%%%%%%%%%%%%%%%%%%%%%%%%%%%%%%%%%%%%%%%%%%%%%%%%

\section{Background}
\label{section:background}

Let us consider a recurrent system with dynamics that can be written as $\vh_t=f(\vh_{t-1}, \vx_t, \boldsymbol\theta)$, where $\vh_t\in\mathbb{R}^n$ is the learner's state vector, $\vx_t\in\mathbb{R}^d$ is the input vector, and $\boldsymbol\theta=\{\mW_1,\dots,\mW_p\}$ is the set of learnable parameters of the recurrence function, where each $\mW\in\boldsymbol\theta$ is a weight matrix. The output is given by $\hat{\vy}_t=g(\vh_t, \vx_t, \boldsymbol\phi)$, where $\hat{\vy}_t\in\mathbb{R}^m$, and $\boldsymbol\phi=\{\mV_1,\dots,\mV_q\}$ is the set of learnable parameters of the output function, where each $\mV\in\boldsymbol\phi$ is a weight matrix. To learn the parameters, we need to compute $\nabla \mathcal{L}_{\mW}, \; \forall \mW \in \boldsymbol\theta$ and $\nabla \mathcal{L}_{\mV}, \; \forall \mV \in \boldsymbol\phi$. Given a target $\vy_t\in \mathbb{R}^m$, the gradient of the loss with respect to $\mW$, treating it as a flattened vector, is given by $\frac{\partial \mathcal{L}(\vy_t, \hat{\vy}_t)}{\partial \mW} =  \left(\frac{\dd \vh_t}{\dd \mW}\right)^\top\frac{\partial \mathcal{L}}{\partial \vh_t}, \quad \forall \mW \in \boldsymbol\theta$.

%%%%%%%%%%%%%%%%%%%%%%%%%%%%%%%%%%%%%%%%%%%%%%%%%%%%%%%%%%%%

\subsection{Backpropagation Through Time (BPTT)}

Note that to obtain $\frac{\dd \vh_t}{\dd \mW}$, we must account for both the direct and indirect effects of changing $\mW$ on $\vh_t$, since $\mW$ influences $\vh_t$ directly and also through every earlier state. Since we can unroll the function $\vh_t=f(f(f(\dots f(\vh_{0}, \vx_{1}, \boldsymbol\theta), \dots \vx_{t-2}, \boldsymbol\theta), \vx_{t-1}, \boldsymbol\theta), \vx_t, \boldsymbol\theta)$, we write the gradient $\frac{\dd \vh_t}{\dd \mW}$ as
\begin{align}
\frac{\dd \vh_t}{\dd \mW} &=  \frac{\partial \vh_t}{\partial \mW} + \frac{\partial \vh_t}{\partial \vh_{t-1}} \frac{\dd \vh_{t-1}}{\dd \mW}, \quad \forall \mW \in \boldsymbol\theta.
\end{align}
We need to keep unrolling further because $\mW$ again affects $\vh_{t-1}$ through two pathways. Let us define $\mS_t \doteq \frac{\dd \vh_t}{\dd \mW}$, $\mJ_t \doteq \frac{\partial \vh_{t}}{\partial \mW}$, and $\mB_t \doteq \frac{\partial \vh_{t}}{\partial \vh_{t-1}}$. We can write the recursive relationship as follows:
\begin{align*}
\mS_t &=  \mJ_t + \mB_t \mS_{t-1} \\
&= \mJ_t + \mB_t \mJ_{t-1} + \mB_t  \mB_{t-1} \mS_{t-2} \\
&= \sum_{j=1}^{t} \left( \prod_{i=j+1}^{t} \mB_{i} \right) \mJ_j. \text{\quad\quad\quad\quad(under the assumption that $\vh_0$ is independent of $\mW$)}\tag{\theequation}\stepcounter{equation}\label{eq:bptt}
\end{align*}
This relationship is utilized in BPTT by efficiently calculating the summation backward without forming or storing $\mS_t$. Calculating this requires storing all previous inputs and states $\vx_i, \vh_i, \forall \{1,\dots,t\}$. In other words, we need to backpropagate the gradient in time, starting from the current time step and going all the way back to the first time step. The computational and memory cost, therefore, grows linearly with the sequence length, since each update must traverse the entire history.

%%%%%%%%%%%%%%%%%%%%%%%%%%%%%%%%%%%%%%%%%%%%%%%%%%%%%%%%%%%%

We can simplify the intensive computation needed by BPTT if we truncate the backpropagation process by going back in time up to time step $t-T$, where $t>T$. This creates the Truncated-Backpropagation Through Time (T-BPTT) gradient \citep{williams1990tbptt}:
\begin{align}
\mS_t  &\approx \sum_{j=t-T}^{t} \left( \prod_{i=j+1}^{t} \mB_{i} \right) \mJ_j, \quad \forall t > T.
\label{eq:tbptt}
\end{align}
T-BPTT reduces the cost of full BPTT by retaining only the most recent $T+1$ inputs and states. Its approximation, however, removes all credit-assignment paths that extend more than $T$ steps into the past, which can make learning myopic. It also requires a finite buffer of past experience, and is therefore not streaming unless $T=1$. We refer to this limiting case as one-step BPTT. Although one-step BPTT propagates gradient credit only one step backward, it can still learn dependencies over longer horizons indirectly, because the recurrent state evolves online and may preserve information from earlier observations.  Algorithm~\ref{alg:tbptt} in Appendix~\ref{appendix:bptt-algorithm} gives the one-step BPTT pseudocode.

\subsection{Real-time Recurrent Learning (RTRL)}
Instead of relying on the rolled-out equation (Eq.\ \ref{eq:bptt} or Eq.\ \ref{eq:tbptt}), we can instead compute $\mS_t=\frac{\dd \vh_t}{\dd \mW}$, known also as the sensitivity matrix, incrementally using its recursive relationship. This process is known as Real-Time Recurrent Learning (RTRL) \citep{williams1989rtrl} given by
\begin{align}
\mS_t &=  \mJ_t + \mB_t \mS_{t-1}.
\label{equ:sensitivity-recurrence}
\end{align}
The quantities $\mJ_t$ and $\mB_t$ depend only on the current input and state, so they can be computed without BPTT; only $\mS_{t-1}$ must be stored to update $\mS_t$ incrementally. Unlike T-BPTT, RTRL computes the exact gradient without approximation, provided the parameters are held fixed. In general, storing and updating $\mS_t$ is computationally expensive and scales poorly with the number of units and parameters, but RTRL becomes practical mainly when the recurrence has an exploitable structure, such as diagonal or sparse dynamics. These savings, however, hold only for a single recurrent layer. When stacked, the input to an upper layer depends on the parameters and past states of the layers below it, so exact sensitivities must also track these cross-layer dependencies and lose their elementwise structure. BPTT does not suffer from this limitation since it never stores $\mS_t$. In practice, RTRL is applied within each layer, treating the output of the layer below as a fixed input, which makes the gradient approximate for the multi-layer networks case \citep{zucchet2023online}. Algorithm~\ref{alg:rtrl} in Appendix~\ref{appendix:rtrl-algorithm} gives the RTRL pseudocode.

%%%%%%%%%%%%%%%%%%%%%%%%%%%%%%%%%%%%%%%

\subsection{State Construction}
\label{section:state-construction}
The recurrent update determines both what enters the state and how existing content is transformed. We focus on recurrences that are linear in the state, the class used by modern linear recurrent and state-space models \citep{gu2021efficiently, orvieto2023resurrecting, gu2024mamba, de2024griffin}. Linearity in the state separates these two roles explicitly and makes stability a property of the transition alone. Any such recurrence can be written as
\begin{align}
    \vh_t = \mA(\vx_t)\vh_{t-1} + \vb(\vx_t).
    \label{eq:affine}
\end{align}
We call $\vb:\mathbb{R}^d\to\mathbb{R}^n$ the \emph{write}: it maps the current input to the new content added to the state. We call $\mA:\mathbb{R}^d\to\mathbb{R}^{n\times n}$ the \emph{transition}: it maps the current input to the matrix that transforms the content already in the state. Both may carry learnable parameters, and we here make no further assumption on either.

\paragraph{Input-independent transition.}
When the transition is a fixed matrix, $\mA(\vx)=\mA$ for every input $\vx$, unrolling Eq.~\ref{eq:affine} from $\vh_0=\mathbf{0}$ gives $\vh_t = \sum_{s=1}^{t} \mA^{t-s}\vb(\vx_s)$.
Each write is therefore transformed according to its age. This supports retention, decay, oscillation, and fixed delays, but the current input cannot change how existing memory is transformed.

Modern linear recurrent and state-space layers exploit this structure directly. For example, LRU and S5 parameterize the fixed transition as a diagonal matrix with complex entries \citep{orvieto2023resurrecting,smith2023simplified}: the magnitude of each entry sets how quickly the corresponding state coordinate decays, and its phase sets how fast it oscillates.
The resulting traces distinguish when an input occurred, but their subsequent evolution is independent of which inputs arrive later.

\paragraph{Input-dependent transition.}
When the transition depends on the current input, unrolling Eq.~\ref{eq:affine} gives $\vh_t = \sum_{s=1}^{t} \left(\prod_{r=s+1}^{t} \mA(\vx_r)\right) \vb(\vx_s)$.
Now, each write $\vb(\vx_s)$ is transformed by the ordered product of all past transitions. A common choice restricts the transition to be diagonal for every input, $\mA(\vx)=\Diag(\va(\vx))$ with $\va:\mathbb{R}^d\to\mathbb{R}^n$ (or $\mathbb{C}^n$). This choice provides useful input-dependent retention and forgetting, but any two such transitions commute: $\mA(\vx_i)\mA(\vx_j)=\mA(\vx_j)\mA(\vx_i)$.
Consequently, exchanging two diagonal transitions does not change how they transform content already in the state. To make that transformation depend on order, the transition family must contain a non-commuting pair,
$\mA(\vx_i)\mA(\vx_j)\neq\mA(\vx_j)\mA(\vx_i)$. Such transitions support computations in which inputs act as ordered operations on memory, including finite-state, permutation, and group tracking \citep{merrill2024illusion}.

Input-dependent diagonal transitions are used by selective state-space and gated linear-recurrent models to control retention from the current input \citep{gu2024mamba,de2024griffin,katsch2023gateloop,qin2024hgrn}. Replacing them with unrestricted dense matrices would permit non-commutativity but would give up the elementwise update and the simple stability constraint that motivates diagonal recurrences: stability would instead depend on controlling the global spectral properties of the transition matrix, while repeated unconstrained matrix products can produce vanishing or exploding dynamics \citep{pascanu2013difficulty,miller2019stable,zucchet2024recurrent}.

\paragraph{Householder transitions.}
A structured transition that does not commute, yet stays cheap and stable, is the generalized Householder transformation \citep{householder1958unitary, grazzi2025unlocking}, $\mA(\vx_t) = \mI-\beta_t\vu_t\vu_t^\top$ with $\lVert\vu_t\rVert=1$.
It leaves the part of the state orthogonal to the axis $\vu_t$ unchanged and scales the part along it by $1-\beta_t$. For $\beta_t\in(0,2]$ it never expands the state, and at $\beta_t=2$ it is a norm-preserving reflection. Two transitions whose axes are neither parallel nor orthogonal do not commute, so the state depends on the order of the inputs.
For example, with $\beta_t=2$, choosing $\vu_t=(\ve_i-\ve_j)/\sqrt{2}$ swaps coordinates $i$ and $j$. By choosing $\vu_t$ according to each successive swap, a one-hot state can track an object through an arbitrary sequence of swaps.
With $\beta_t\in(0,1)$, this is the transition of DeltaNet \citep{schlag2021linear, yang2024parallelizing}; extending $\beta_t$ to $(0,2)$ lets linear RNNs track parity and permutations \citep{grazzi2025unlocking, siems2025deltaproduct}.

%%%%%%%%%%%%%%%%%%%%%%%%%%%%%%%%%%%%%%%%%%%%%%%%

\section{Streaming Recurrent Memory}
\label{section:srm}

We want a recurrent memory that can be trained with one-step BPTT and still compose inputs in an order-sensitive way.
We design it around four desiderata. (D1) \emph{Reliable credit-free write}: as one-step BPTT feedback cannot reach how earlier inputs were written, each input should enter the state in a form that can be retrieved later, without requiring the writing mechanism to be learned. (D2) \emph{Cheap computation}: the recurrence should be structured such that its parameters can be learned cheaply. (D3) \emph{Order sensitivity}: part of the state should evolve under input-dependent transitions that do not commute. (D4) \emph{Stability}: repeated transitions should stay bounded by construction.

SRM specializes Eq.~\ref{eq:affine} into two states, a matrix memory $\mH_t\in\mathbb{R}^{d_k\times d_v}$ with key width $d_k$ and value width $d_v$, and a tracking vector $\vz_t\in\mathbb{R}^{G}$:

\begin{align}
    \mH_t = \boldsymbol\Lambda(\vx_t)\,\mH_{t-1} + \vk_t\,\vv_t^\top,
    \qquad
    \vz_t = \left(\mI-\beta_t\,\vu_t\vu_t^\top\right)\vz_{t-1},
    \qquad
    \vr_t = \begin{bmatrix}\vq_t^\top\mH_t\\ \vz_t\end{bmatrix}.
    \label{eq:srm}
\end{align}

The memory decays each row by the retention $\boldsymbol\Lambda(\vx_t)=\diag\big(\exp(-\exp(\mW_\lambda\vx_t+\vb_\lambda))\big)$ and stores the value $\vv_t=\mV_x\vx_t+\mV_p\vx_{t-1}$ under the unit-norm key $\vk_t=\bar\vk_t/\lVert\bar\vk_t\rVert$, with $\bar\vk_t=\mK_x\vx_t+\mK_p\vx_{t-1}+\mK_c\vc_t$, where $\vc_t$ is a fixed oscillatory clock. These key and value projections incorporate causal 1D convolutions with a kernel size of two. The memory is read with the query $\vq_t=\mW_{q,x}\vx_t+\mW_{q,c}\vc_t$. The tracking vector is reflected along the axis $\vu_t=\mU\vx_t/\lVert\mU\vx_t\rVert$ with strength $\beta_t=2\sigma(\vw_\beta^\top\vx_t+b_\beta)$. The maps $\mK_x$, $\mK_p$, $\mK_c$, $\mV_x$, $\mV_p$, and $\mU$ are fixed random semi-orthogonal projections, so the only learned parameters are the query maps ($\mW_{q,x}, \mW_{q,c}$), the retention $(\mW_\lambda,\vb_\lambda)$, and the strength $(\vw_\beta,b_\beta)$. Neither state enters the other's update, and $\vr_t\in\mathbb{R}^{d_v+G}$ is passed to the output function $g$. In practice, SRM uses $N_\text{h}$ read heads with separate query maps and concatenates their reads, so its output has width $N_\text{h}d_v+G$.

\subsection{Fixed content--time write}
\label{section:srm-write}

Under one-step BPTT, $\vh_{t-1}$ is treated as a constant, so feedback at time $t$ has no gradient path to a write made at any earlier step.
A learned write would therefore be trained only by the feedback available when it writes, which cannot indicate what a later step will need. SRM therefore fixes the write and learns the read, partly achieving D1. The key combines content, the current and previous input, with time, a clock whose rotors cycle at geometrically spaced frequencies (see Appendix~\ref{appendix:srm-clock} for more details), so a stored value can later be found by what preceded it or by when it occurred. The value carries both current and previous inputs, so a retrieval returns the pair. The fixed maps are semi-orthogonal, which keeps distinct inputs at distinct addresses, and the key is normalized, so every write has the same magnitude.

Key--value outer products are the classical linear associative memory \citep{willshaw1969nonholographic, anderson1972simple, kohonen1972correlation}, revisited as fast weights \citep{schmidhuber1992learning, ba2016using, schlag2021linear} and linear attention \citep{katharopoulos2020linear}. Fixing the input maps and learning only a readout is the idea of reservoir computing \citep{jaeger2001echo, maass2002real}, which SRM applies to the write alone. Keys that see the previous input are related to short convolutions in linear-recurrent layers \citep{fu2023hungry, team2025kimi} and to previous-token heads in Transformers \citep{olsson2022context}, and the clock follows sinusoidal and rotary position encodings \citep{vaswani2017attention, su2024roformer}.

\subsection{Learned associative read}
\label{section:srm-read}

For a memory $\sum_i\vk_i\vv_i^\top$, the read $\vq_t^\top\sum_i\vk_i\vv_i^\top=\sum_i\langle\vq_t,\vk_i\rangle\vv_i$ returns the values whose keys match the query, plus interference from other keys that is small when keys are nearly orthogonal. The query has a content part and a time part, matching the key, so retrieval can select by content, by time, or both. We initialize $\mW_{q,x}$ as a semi-orthogonal map scaled by $1/\sqrt{d}$ and $\mW_{q,c}$ at zero, so time addressing is learned. Because the query acts only on the current memory, its one-step BPTT gradient is exact (D1).

This is the read of linear attention and fast weights \citep{katharopoulos2020linear, schlag2021linear} and of gated linear-attention and state-space models \citep{sun2023retentive, yang2024gla, dao2024transformers, beck2024xlstm}, whose capacity limits are well studied \citep{arora2024zoology, arora2024simple, okpekpe2025revisiting}. Delta-rule models reduce interference by learning to erase before writing \citep{schlag2021linear, yang2024parallelizing, yang2025gated}. SRM relies on nearly orthogonal fixed keys instead, because learning what to erase would need gradient from later feedback.

\subsection{Diagonal decaying retention}
\label{section:srm-transition}

The retention should control how long memory persists without moving what is stored. Row $s$ of the memory is scaled by $\lambda_{t,s}=\exp(-\exp(\vw_{\lambda,s}^\top\vx_t+b_{\lambda,s}))\in(0,1)$, so every step is contractive (D4) and exact RTRL sensitivities are elementwise and thus cheap to compute (D2). Unrolling gives $\mH_t=\sum_{\tau\le t}\big(\prod_{s=\tau+1}^{t}\boldsymbol\Lambda(\vx_s)\big)\vk_\tau\vv_\tau^\top$: a stored key keeps its coordinates up to a decay, so the query that matched it when it was written still matches it later. A transition that rotated or mixed key coordinates would move stored addresses in ways a query cannot undo (see Section~\ref{subsec:ablation}). At initialization, $\vw_{\lambda,s}=\mathbf{0}$ and $b_{\lambda,s}=\log(-\log r_s)$ with $r_s$ drawn from the LRU ring distribution \citep{orvieto2023resurrecting}, so the memory starts as leaky integrators over a range of time scales.
Under one-step BPTT, the retention and the tracking strength are the only learned parameters whose gradient is used: they receive credit for their effect on the current step, but not for how their earlier values shaped the state being read.

Input-dependent diagonal gates, descending from LSTM and GRU gating \citep{hochreiter1997long, cho2014learning}, control retention in selective state-space, gated linear-recurrent, and gated linear-attention models \citep{gu2024mamba, de2024griffin, katsch2023gateloop, qin2024hgrn, yang2024gla, feng2024were}; SRM's double-exponential parameterization follows LRU \citep{orvieto2023resurrecting}. Such transitions commute, which limits order-sensitive computation \citep{merrill2024illusion}.

\subsection{Householder tracking channel}
\label{section:srm-track}

Because the retention commutes, the memory cannot follow a state that evolves under non-commuting operations, such as which cup hides an object after a sequence of swaps. The tracking vector does this with the Householder transition (D3), which never expands $\vz_t$ because $\beta_t\in(0,2)$ (D4).
The axis map $\mU$ is fixed at a random semi-orthogonal initialization. The strength is initialized such that $\beta_t=1.9$ ($\vw_\beta=\mathbf{0}$, $b_\beta=2\operatorname{arctanh}(0.9)$).

DeltaNet and its gated variants \citep{schlag2021linear, yang2024parallelizing, yang2025gated, team2025kimi} apply this transformation to the memory itself, with the written key as axis. SRM applies it to a separate small state with a fixed axis map, so tracking never moves the addresses of the associative memory.

%%%%%%%%%%%%%%%%%%%%%%%%%%%%%%%%%%%%%%%%%%%%%%%%%%%%%%%%%%%

\section{Experiments}

We first use a diagnostic task suite evaluate the units at initialization, the reliability of one-step BPTT gradients, and streaming learning performance. We then test SRM on Multi-Query Associative Recall \citep{arora2024zoology}, streaming byte-level language modeling on TinyStories \citep{eldan2023tinystories}, and control from delayed reward  in an associative T-Maze, and finally ablate its components.

We choose four recurrent baselines for their varying differences from SRM. Linear RTU \citep{elelimy2024realtime} uses an input-independent complex-diagonal state transition. eLSTM \citep{irie2024exploring} uses input-dependent gates and its recurrence is nonlinear in the state. GLA \citep{yang2024gla}, like SRM, is a linear attention unit with input-dependent diagonal gating, providing a streamlined architectural baseline. KDA \citep{team2025kimi} applies a generalized Householder update directly to the memory channel and shares SRM’s use of a 1D convolution.
All learners operate under the streaming setting, that is, without replay, batches, or retained context. Performance is evaluated online: accuracy, success rate, or loss is measured on each sample before the learner updates on it. We present the key results here and give the full details in Appendix~\ref{appendix:exp-details}.

\subsection{Diagnostics}
\label{sec:diagnostics}

We introduce a task suite (Table~\ref{tab:task-suite0} and Appendix~\ref{appendix:task-suite}) developed to evaluate SRM along the design desiderata. The suite comprises ten tasks, each involving a different type of information, to provide a holistic evaluation of the ability to store and retain incoming information (D1). The suite covers tasks that test the capabilities to recall information by content or time, as well as to track counts, orders, and states. Two of the tasks (\texttt{Orient} and \texttt{Shell}) concern tracking non-commuting operations (D3).

\begin{table}[ht]
\centering
\scriptsize
\renewcommand{\arraystretch}{1.2}
\begin{tabular}{lp{0.7\linewidth}}
\toprule
\textbf{Task} & \textbf{Target at query time} \\
\midrule
\texttt{Presence} & Whether a queried symbol appeared in the sequence. \\
\texttt{Successor} & The symbol that appeared right after the queried symbol. \\
\texttt{Predecessor} & The symbol that appeared right before the queried symbol. \\
\texttt{Count} & The number of occurrences of the queried symbol. \\
\texttt{Order} & Whether the first queried symbol appeared before the second in the presentation. \\
\texttt{Position} & The symbol presented at a queried position. \\
\texttt{Parity} & Addition modulo 2---tracking under the symmetric group $S_2$. \\
\texttt{Orient} & Composition of rotations and reflections---tracking under the dihedral group $D_4$.\\
\texttt{Shell} & Composition of cup permutations---tracking under the symmetric group $S_3$. \\
\texttt{Mod3} & Addition modulo 3---accumulation in the cyclic group $\mathbb{Z}_3$. \\
\bottomrule
\end{tabular}
\caption{The task suite. Each task tests how well a unit encodes and retains one kind of information. Each sequence consists of 5 to 8 task-dependent input symbols, followed by a sequence of waiting symbols. At the end of the sequence, the learner is prompted to give an answer, at which point feedback is provided.}
\label{tab:task-suite0}
\end{table}

\subsubsection{Memory Capabilities without Learning}
\label{subsec:init}

We evaluate how well each unit can encode and retain information before any learning happens (D1). We unfold the units through a fixed set of examples, then continue the unfolding through a waiting period to test how well information is retained. We record the recurrent states after $0$, $100$, $200$, and $500$ of these waiting timesteps.
At each checkpoint, we fit a fresh ridge probe on the states to predict the target (Figure~\ref{fig:suite_init}); details in Appendix~\ref{appendix:init-exp}.

\begin{figure}[ht]
    \centering
    \includegraphics[width=0.9\linewidth]{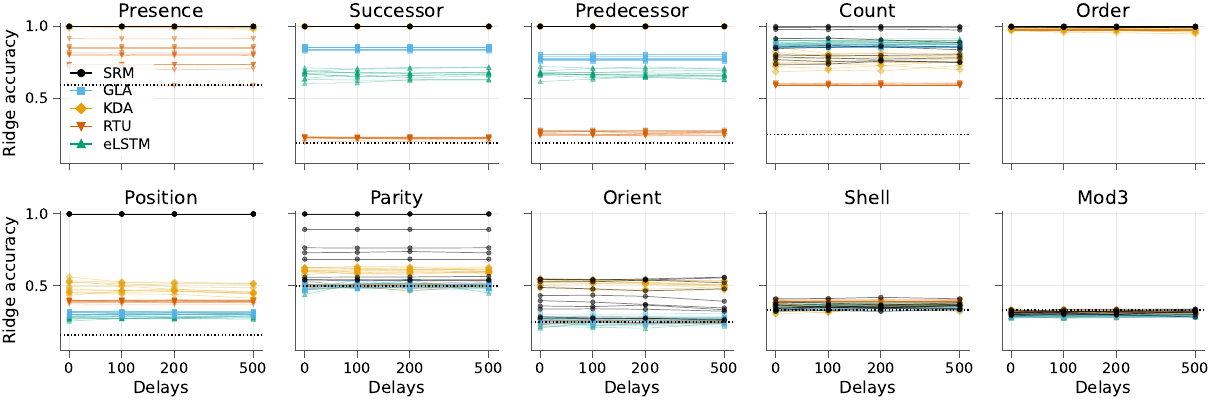}
    \caption{Accuracy of ridge-probe over recurrent states unfolded to delay checkpoints, without training. Lines connect points from each of the 10 random initializations. Dotted lines indicate chance-level accuracy.}
    \label{fig:suite_init}
\end{figure}

SRM can encode and retain information well in most tasks. The previous-input projections, $\mK_p\vx_{t-1}$ and $\mV_p\vx_{t-1}$, enable SRM to encode \texttt{Successor} and \texttt{Predecessor} well. The clock term, $\mK_c\vc_t$, is essential in \texttt{Position}. The probe falls short in \texttt{Orient}, and fails in \texttt{Shell} and \texttt{Mod3} across all units, suggesting that the information is not present at initialization. Overall, compared to  baseline units, SRM exhibits a greater capacity to store and retain information at initialization.

\subsubsection{Reliability of one-step BPTT in learning SRM parameters}
\label{subsec:gradients}
We compare the reliability of learning with one-step BPTT relative to full BPTT. At time $T$, the latter computes the gradients for each parameter by accounting for the contributions of all steps in the past: $\vg_\mathrm{full} = \sum_{t=0}^{T} \vg_t$. On the other hand, one-step BPTT only uses the most recent step and omits the contributions from all earlier steps: $\vg_\mathrm{one\_step} = \vg_T = \vg_\mathrm{full} - \vg_\mathrm{earlier\_steps}$. Learning with one-step BPTT becomes myopic unless the omitted $\vg_\mathrm{earlier\_steps}$ is captured by $\vg_\mathrm{one\_step}$.
We can measure the direction alignment between $\vg_\mathrm{earlier\_steps}$ and $\vg_\mathrm{one\_step}$, for a parameter, with cosine similarity, and denote this alignment score as $\rho$. Negative $\rho$ indicates that one-step BPTT degrades the parameters beyond one-step memory. The learning becomes unreliable if misalignment occurs frequently. We study how this alignment varies across tasks and changes as the loss evolves, for SRM.

\begin{figure}[ht]
    \centering
    \includegraphics[width=0.99\linewidth]{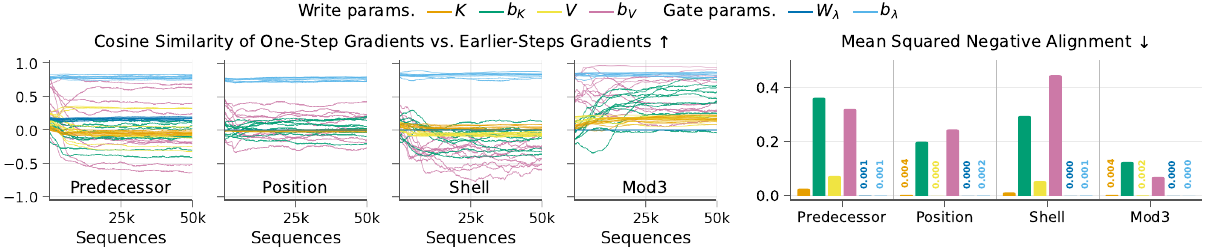}
    \caption{(Left) Alignment score ($-1$ to $1$; higher is better) throughout readout training, for all of $10$ SRM initializations. (Right) Squared negative alignment ($0$ to $1$; lower is better), averaged over all steps and runs.}
    \label{fig:alignment_summary}
\end{figure}

For each task, we train the readout parameters to allow the loss to evolve in a controlled manner. While we keep all non-readout parameters fixed throughout this procedure, we differentiate the loss with respect to selected parameters to obtain their alignment scores. We compute $\rho$ for the key projection matrix $\mK$ (which denotes the concatenation of $\mK_x$, $\mK_p$, and $\mK_c$), the value projection matrix $\mV$ (concatenation of $\mV_x$ and $\mV_p$), and the gate's parameters $\mW_\lambda$, $\vb_\lambda$. We additionally track $\vb_{\mK}$ and $\vb_{\mV}$, two bias terms for the key and the value that are functionally fixed at 0 by SRM. Figure~\ref{fig:alignment_summary} shows a representative subset of results. We also compute the squared negative alignment, $\min(0, \rho)^2$, and average it over all training steps and runs. Full details are in Appendix~\ref{appendix:gradient-exp}.

The results indicate that the learning of the gate parameters is, at the very least, innocuous. Across all tasks and readout conditions, $\rho_{\mW_\lambda}$ rarely ever falls below $0$. $\rho_{\vb_\lambda}$ stays consistently high, indicating that the retention strength is well learnable with one-step BPTT. For the write parameters, $\rho_{\mK}$ and $\rho_{\mV}$ remain near 0 throughout but occasionally become negative. Even though $\rho_{\vb_{\mK}}$ and $\rho_{\vb_{\mV}}$ are typically positive in \texttt{Mod3}, they exhibit high run-to-run variance in all tasks. In many runs, they suffer from negative alignment, sometimes severely, and do not improve with the condition of the readout. Overall, the results suggest that learning the write parameters with one-step BPTT is not a reliable process.

\subsubsection{Memory Capabilities with Streaming Learning} 
\label{subsec:train}

\begin{figure}[ht]
    \centering
    \includegraphics[width=0.9\linewidth]{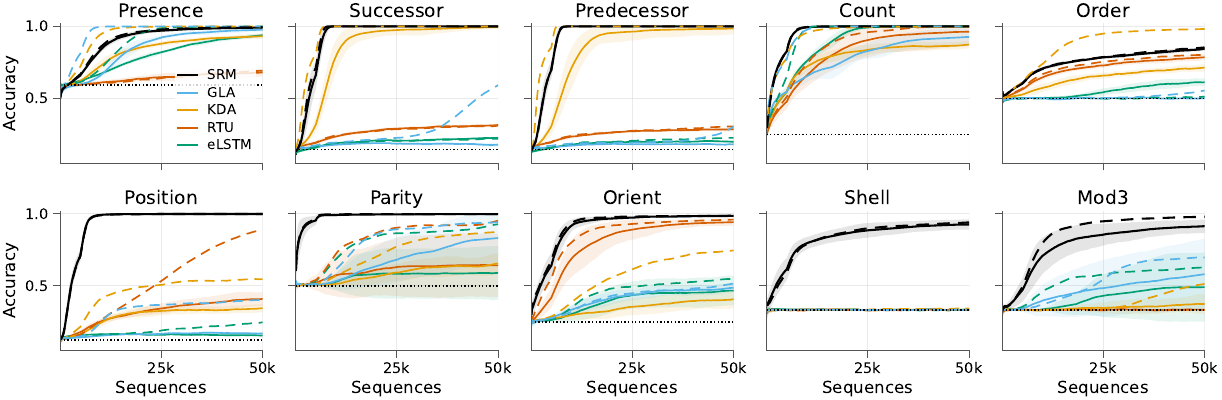}
    \caption{Training accuracy on the task suite under one-step BPTT (solid lines) and RTRL (dashed lines), averaged over $10$ runs. Shading shows standard deviation for one-step BPTT. Dotted lines indicate chance.}
    \label{fig:suite_train}
\end{figure}

We examine whether the observed nascent memory capabilities at initialization translate to realized capabilities, under one-step BPTT training (solid lines in Figure~\ref{fig:suite_train}). On every task where untrained SRM has high probe accuracy relative to the other units, SRM trained under one-step BPTT also obtains an online accuracy higher than other units. This pattern is clear in all of the first seven tasks. In \texttt{Orient}, \texttt{Shell} and \texttt{Mod3}, where the ridge probe struggles, SRM performs well with training nevertheless. Overall, no unit outperforms SRM, across every task, when trained under one-step BPTT.

We also report RTRL results (dashed lines in Figure~\ref{fig:suite_train}). With one recurrent layer, RTRL updates are exact, and serve as an approximate upper bound performance on unit's performance given its capacity. Compared to baseline units, SRM under one-step BPTT closely matches its RTRL performance, across all tasks, suggesting that its design is particularly effective for one-step BPTT, which also better satisfies (D2) than RTRL (Appendix~\ref{appendix:rtrl-empirical}). From here on, we focus on one-step BPTT.

\subsection{Streaming Learning on Multi-Query Associative Recall}
We evaluate SRM on the Multi-Query Associative Recall (MQAR) task \citep{arora2024zoology}, where the learner must recall multiple values associated with their corresponding keys. The demand to store multiple independent bindings simultaneously challenges associative memory with increased interference among bindings. Table~\ref{tab:tasks} (and Figure~\ref{fig:tasks}a) reports the late-training accuracy on MQAR at three model dimensions across two task settings. Across tested model capacities, SRM outperforms all baselines, including KDA, which also uses causal 1D convolutions effective for this task. These results show that SRM's gains extend beyond the task suite to established benchmarks for context-aware retention and associative binding.

\subsection{Streaming Byte-Level Language Modeling}
\label{subsec:language}

Next, we train language models one byte at a time on the TinyStories training set \citep{eldan2023tinystories}. The learner reads the stories in dataset order with batch size one and resets its recurrent state at every story boundary. At each byte, it predicts the next byte, incurs the cross-entropy loss, and makes one update before the next byte arrives. Each model creates an embedding for it, passes it through four stacked residual blocks of recurrent units (see Appendix \ref{appendix:lm-exp}), and then predicts the next byte with a linear head. All models are trained with one-step BPTT for $100$M bytes, about $5\%$ of the roughly $2$B-byte corpus (see Figure~\ref{fig:tasks}).

\begin{table}[ht]
\centering
\scriptsize
\renewcommand{\arraystretch}{1.15}
\begin{tabular}{lcccccccc}
\toprule
& \multicolumn{6}{c}{\textbf{MQAR accuracy (\%)} $\uparrow$} & \multicolumn{2}{c}{\textbf{TinyStories loss} $\downarrow$} \\
\cmidrule(lr){2-7} \cmidrule(lr){8-9}
& \multicolumn{3}{c}{Sequence length 64} & \multicolumn{3}{c}{Sequence length 128} & \multicolumn{2}{c}{Bytes seen} \\
\cmidrule(lr){2-4} \cmidrule(lr){5-7} \cmidrule(lr){8-9}
\textbf{Unit} & Small & Medium & Large & Small & Medium & Large & $10$M & $20$M \\
\midrule
GLA   & 11.1 & 17.6 & 23.3 & 1.3 & 8.3 & 11.3 & $1.193 \pm 0.029$ & $1.192 \pm 0.048$ \\
KDA   & 5.6 & 20.1 & 92.4 & 2.7 & 5.2 & 62.3 & $1.076 \pm 0.006$ & $0.994 \pm 0.009$ \\
RTU   & 0.8 & 0.8 & 14.0 & 0.9 & 0.7 & 2.3 & $0.942 \pm 0.002$ & $0.892 \pm 0.002$ \\
eLSTM & 7.3 & 13.8 & 22.6 & 2.3 & 1.8 & 9.6 & $1.307 \pm 0.017$ & $1.130 \pm 0.015$ \\
SRM   & \textbf{22.5} & \textbf{77.3} & \textbf{99.1} & \textbf{12.4} & \textbf{51.6} & \textbf{96.9} & $\mathbf{0.855 \pm 0.003}$ & $\mathbf{0.822 \pm 0.003}$ \\
\bottomrule
\end{tabular}
\caption{MQAR: accuracy over the final $10\%$ of training sequences at three FLOPs-matched model dimensions (Table~\ref{tab:mqar-capacity}), averaged over $10$ runs. TinyStories: online next-byte loss averaged over the $1$M bytes preceding the $10$M- and $20$M-byte marks, reported as mean $\pm$ standard deviation over three independent runs; all language models have around $13$M parameters. Best value per column is shown in bold.}
\label{tab:tasks}
\end{table}

Table~\ref{tab:tasks}(b) reports the online loss after $10$M and $20$M bytes, and Figure~\ref{fig:tasks}(b) in the Appendix shows the full learning curves. Among the baselines, RTU reaches the lowest online loss, followed by KDA and eLSTM, while GLA remains the highest throughout.
SRM has the lowest online loss of all units from the first few million bytes onward and reaches around $0.82$, below RTU, which achieves about $0.89$. At this point, SRM can already produce locally coherent TinyStories-like text: its samples follow the typical story structure and are mostly grammatical within a sentence, although characters, pronouns, and events are not kept consistent across sentences (see Appendix~\ref{appendix:lm-sample} for a generation sample). We also found that, in this regime, the optimizer matters as much as the unit: with RMSprop \citep{tieleman2012lecture}, the online loss can stop improving after a few million bytes and then drifts upward, whereas StreamingOptimizer \citep{elsayed2026streaming} keeps decreasing (see Appendix~\ref{appendix:lm-optimizer}). Furthermore, we study performance scaling with parameter count in Appendix~\ref{appendix:lm-scaling}, showing it scales under streaming learning.

\subsection{Memory-Dependent Control with Streaming Reinforcement Learning}
\label{subsec:assoc-tmaze}

Finally, we ask whether SRM's binding and retrieval survive when the only feedback is a delayed scalar reward. In an associative variant of the T-Maze \citep{bakker2001reinforcement}, the agent observes $K$ key--value pairs one token at a time, each value being \texttt{left} or \texttt{right}, waits through a delay, and is then shown one of the keys at a junction; turning toward that key's value gives a reward of $+1$ and the other turn $-1$. With $K=1$ this is the standard memory T-Maze, and with $K>1$ the agent must bind each value to its key, a reinforcement-learning counterpart of \texttt{Successor}. We used the Stream-AC algorithm \citep{elsayed2026streaming} that updates at every step. We compare SRM against RTU  \citep{elelimy2024realtime} and report the average online success rate and the final success rate over ten independent runs (see Appendix~\ref{appendix:assoc-tmaze} for more details).

\begin{figure}[H]
    \centering
    \includegraphics[width=0.95\linewidth]{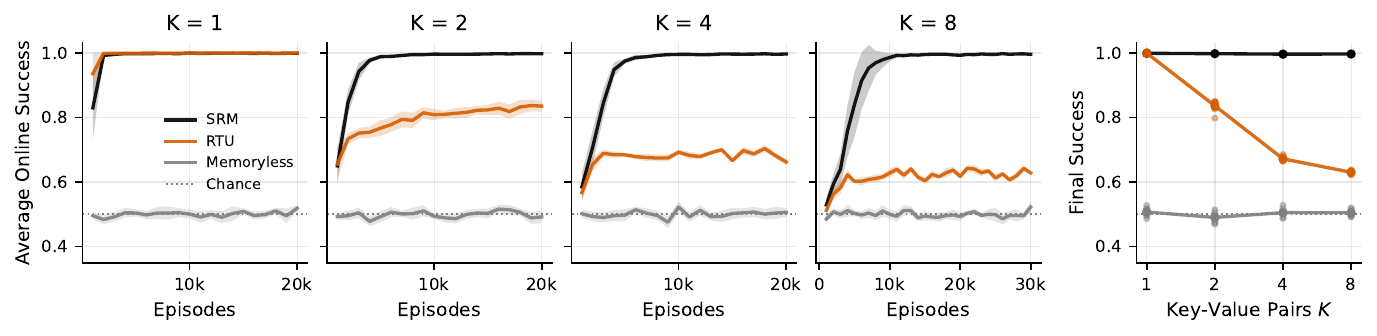}
    \caption{Associative T-Maze with $K$ key--value pairs. Left four panels: online success rate during training averaged over $10$ runs with shaded area representing one standard error. Dotted line is chance. Rightmost panel shows online success over the final $10\%$ of episodes (dots are individual seeds).}
    \label{fig:assoc-tmaze}
\end{figure}

Figure~\ref{fig:assoc-tmaze} shows that both units solve $K=1$, but they have different performance once values must be bound to keys: SRM's final online success stays above $0.99$ for $K=2$, $4$, and $8$, whereas RTU plateaus at $0.84$, $0.67$, and $0.63$. We attribute the gap to the write rather than to credit assignment. SRM's key includes the previous input, so each value is already stored under an address that identifies its key, and the reward only has to train the read.

\subsection{Ablation studies}
\label{subsec:ablation}

\begin{figure}[ht]
    \centering
    \includegraphics[width=0.99\linewidth]{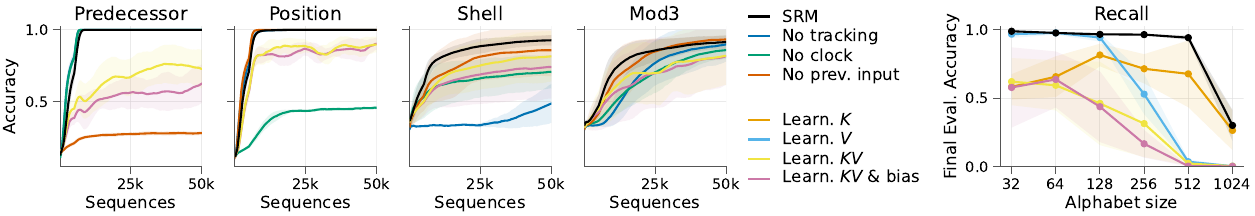}
    \caption{(Left) Ablation of SRM's components: tracking channel, clock, and previous-input projections. Results show the average and standard deviation of 10 runs. (Right) SRM's recall accuracy degrades when the write weights ($\mK$ and $\mV$) and their bias terms are made learnable, across various input alphabet sizes.}
    \label{fig:ablation-toy}
\end{figure}

The ablation results in Figure~\ref{fig:ablation-toy} (Left) suggest that different tasks rely on different components of SRM, which together form a strong overall architecture. The previous-input projections ($\mK_p\vx_{t-1}$ and $\mV_p\vx_{t-1}$) are essential in \texttt{Predecessor} (and also \texttt{Successor}). The clock is essential in \texttt{Position}, and also handy in \texttt{Shell} and \texttt{Mod3}. The tracking channel is necessary for SRM to perform non-commutative tracking reliably.

We study the role of freezing write parameters in SRM through a synthetic recall task in Figure~\ref{fig:ablation-toy} (Right). Training sequences are constructed using symbols randomly sampled from a discrete alphabet. The learner must recall the first-seen symbol at the end of the sequence. Enlarging the alphabet while keeping the model capacity fixed creates a bottleneck in the model's ability to write distinct symbol representations into its recurrent state. The overlapping representation of distinct input symbols consequently leads to greater interference from other keys during retrieval. 

Since learning write parameters with one-step BPTT is unreliable, SRM freezes them at a semi-orthogonal initialization. Compared with variants whose write weights ($\mK$ and $\mV$) and bias terms are learnable, frozen-write SRM achieves the highest recall accuracy and remains robust as the alphabet grows up to $512$. We also overlay the learning curves of SRM variants with learnable write parameters in Figure~\ref{fig:ablation-toy} (Left) to showcase the role of freezing write parameters beyond the recall task.

Next, we investigate the design of SRM's tracking channel. SRM applies its Householder transition to a separate vector $\vz_t$ rather than to the memory $\mH_t$. We test this choice by merging the two states in two ways, with all parameters frozen at initialization so that the results reflect what the state update preserves rather than what training allows (Figure~\ref{fig:hh-separation}; details in Appendix~\ref{appendix:hh-separation}).

Reflecting the memory moves stored keys. Applying the reflection to the memory as well, with a constant strength $\beta$, gives $\mH_t=(\mI-\beta\vu_t\vu_t^\top)\boldsymbol\Lambda(\vx_t)\mH_{t-1}+\vk_t\vv_t^\top$, as in delta-rule models \citep{yang2025gated, team2025kimi}. This update moves every stored key with each later input, so the query that matched it at write time no longer does. The measured overlap with the written position decreases with the number of subsequent inputs (Figure~\ref{fig:hh-separation}a). With $\beta=1.9$, retrieval after $32$ and $64$ inputs falls to $0.6$ and $0.3$, against $0.95$ and $0.9$ for SRM (Figure~\ref{fig:hh-separation}b).

\begin{figure}[ht]
    \centering
    \includegraphics[width=0.95\linewidth]{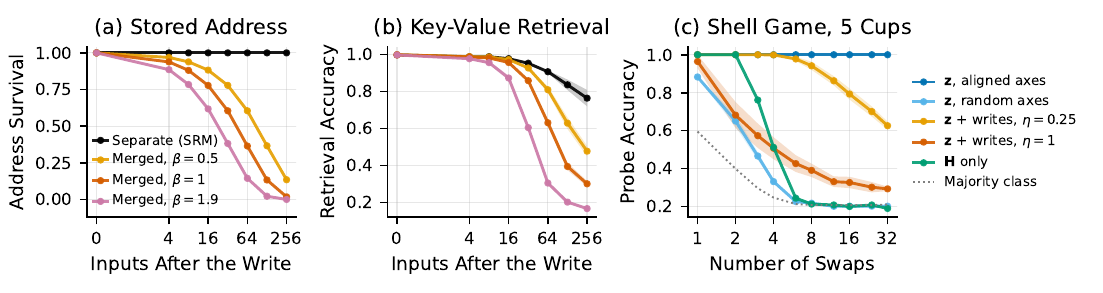}
    \caption{Merging the tracking channel and the memory, with SRM at initialization. (a) Fraction of a stored key left at its written position after number of inputs. (b) Key--value retrieval accuracy; the dotted line is chance. (c) Linear-probe accuracy for the ball position in a five-cup shell game.}
    \label{fig:hh-separation}
\end{figure}

Conversely, writing into the tracking vector corrupts it. In a five-cup shell game where each swap of cups $i$ and $j$ reflects $\vz_t$ along $\ve_i-\ve_j$ with $\beta=2$, $\vz_t$ is exactly the one-hot position of the ball, and a linear probe is perfect at every length. We then add a write to this update, $\vz_t=(\mI-\beta\vu_t\vu_t^\top)\vz_{t-1}+\eta\vw_t$, where $\vw_t=\mW_w\vx_t/\lVert\mW_w\vx_t\rVert$ is the input under a fixed random map $\mW_w$, normalized to unit length, and $\eta$ sets the size of the write relative to the unit-norm $\vz_t$. This lowers accuracy after $32$ swaps to $0.6$ for $\eta=0.25$ and $0.3$ for $\eta=1$ (Figure~\ref{fig:hh-separation}c). The memory alone is near chance after six swaps, because its diagonal transitions commute. 
SRM's random-axis tracking vector is also near chance after $6$ swaps, so its gains on state tracking (Figure~\ref{fig:suite_train}) come from training.

%%%%%%%%%%%%%%%%%%%%%%%%%%%%%%%%%%%%%%%%%%%%%%%%%%%%%%%%%%%

\section{Related Work}
\label{section:related-works}

\textbf{Streaming learning.} Most deep learning methods assume access to the entire dataset, whereas real-world applications rely on a continuous stream of experience. A few supervised methods operate directly under the streaming setting \citep{hayes2019memory, saran2023streaming, hayes2022online}, and related efforts have adapted reinforcement learning \citep{elsayed2026streaming, vasan2024deep}, language modeling \citep{goyal2009streaming}, and video prediction \citep{carreira2024learning, qian2024streaming, han2025learning} to the same regime. These works establish that streaming learning is feasible, but they largely consider fully observable settings where memory is not needed. Memory under streaming learning is what we study here.

\textbf{Recurrent and state-space sequence models.} Linear recurrent and state-space models have re-emerged as efficient alternatives to attention, trading quadratic context for a bounded recurrent state \citep{orvieto2023resurrecting, gu2024mamba, de2024griffin, beck2024xlstm}. Much of their efficiency comes from diagonal or complex-diagonal transitions and parallel-scan training \citep{katharopoulos2020linear, sun2023retentive, katsch2023gateloop, qin2024hgrn, feng2024were}. The same diagonal structure makes their transitions commute, which limits order-sensitive computation \citep{merrill2024illusion}, and these models are usually trained by backpropagation through time over windows with state resets at segment boundaries \citep{dai2019transformer, hutchins2022block}. SRM departs on both axes: it keeps a commuting diagonal retention for its associative memory but adds a separate non-commuting tracking channel, and it is designed to be trained in a streaming manner without resetting or storing contexts.

\textbf{Online credit assignment.} Real-time recurrent learning (RTRL) \citep{williams1989rtrl} computes exact gradients forward in time without storing the stream. Its cost has led to many approximations (e.g., \citealt{tallec2018unbiased, zucchet2023online, elelimy2024realtime}). SRM is designed for the streaming setting. Its write has no parameters and its tracking channel uses a fixed axis map. Exact RTRL is cheap for both components of SRM: the retention sensitivity is elementwise across memory rows, and the tracking-strength sensitivity has only $G\times(d+1)$ entries.

%%%%%%%%%%%%%%%%%%%%%%%%%%%%%%%%%%%%%%%%%%%%%%%%%%%%%%%%%%%

 \section{Limitations and Conclusion}
\label{section:conclusion}

SRM's fixed write bounds its capacity: stored values are separated only by near-orthogonal fixed keys, so interference grows with the number of items held, and SRM cannot learn what to erase. The tracking channel learns only its strength, so what it can track depends on the input representation. Our experiments are small in scale: diagnostic tasks with short sequences, a simple reinforcement-learning task, and small language models by current standards. We did not explore larger language models because of the compute cost of streaming learning: updating after every byte at batch size one limits parallelism across steps and makes it harder to exploit GPUs’ parallel processing capabilities. Hardware optimized for low-batch autoregressive decoding may help address this constraint, provided it also efficiently supports gradient updates.

We studied recurrent memory under streaming learning, with one-step BPTT, where a useful memory must keep past information retrievable while the feedback at each step can only shape how that state is used. Because this feedback cannot reach past writes, SRM fixes its content--time write and learns only how to read, and it places order-sensitive computation in a separate Householder tracking channel so that tracking never disturbs stored addresses. Across diagnostics, controlled tasks, and reinforcement learning, SRM learns memory-dependent behavior under one-step BPTT that the baselines often do not: it retains information that baselines lose, and in the associative T-Maze it binds and retrieves key--value pairs from delayed reward. It also reaches the lowest online loss among all units as a streaming byte-level language model. These results suggest that when credit cannot be carried back through time, how the state is written matters as much as how it is learned.

%%%%%%%%%%%%%%%%%%%%%%%%%%%%%%%%%%%%%%%%%%%%%%%%%%%%%%%

\bibliography{main}
\bibliographystyle{tmlr}

\clearpage
\appendix

\section{The One-step BPTT Algorithm}
\label{appendix:bptt-algorithm}
\begin{algorithm}[H]
\small
\caption{The One-step BPTT Algorithm}
\label{alg:tbptt}
\begin{algorithmic}[1]
\State \textbf{Require}: Recurrent function \(f_{\boldsymbol\theta}\) with parameters \(\boldsymbol\theta\)
\State \textbf{Require}: Output function \(g_{\boldsymbol\phi}\) with parameters \(\boldsymbol\phi\)
\State \textbf{Require}: Data stream \(\mathcal{D}\), step size \(\alpha\)
\State \textbf{Initialize}: \(\boldsymbol\theta,\boldsymbol\phi\), hidden state \(\mathbf{h}_0 \gets \mathbf{0}\)
\For{$t=1\dots \infty$}
    \State $\delta_{\mW} \gets \mathbf{0}, \forall \mW \in \boldsymbol\theta, \quad
    \delta_{\mV} \gets \mathbf{0}, \forall \mV \in \boldsymbol\phi$
    \State $\vx_t,\vy_t, \texttt{reset} \leftarrow \mathcal{D}$
    \If{\texttt{reset}}
        \State $\vh_{t-1} \leftarrow \boldsymbol{0}$
    \EndIf
    \State $\vh_t \gets f_{\boldsymbol\theta}(\vh_{t-1}, \vx_t)$ \Comment{Forward pass}
    \State $\hat{\vy}_t \gets g_{\boldsymbol\phi}(\vh_t, \vx_t)$
    \State Compute $\mathcal{L}_t = \mathcal{L}(\hat{\vy}_t, \vy_t)$
    \State $\delta_{\mV} \gets \left(\frac{\partial \hat{\vy}_t}{\partial \mV}\right)^\top \frac{\partial \mathcal{L}_t}{\partial \hat{\vy}_t}, \quad \mV \in \boldsymbol\phi$ \Comment{Gradient for \( \boldsymbol\phi \)}
    \State $\delta_{\vh} \gets \left(\frac{\partial \hat{\vy}_t}{\partial \vh_t}\right)^\top\frac{\partial \mathcal{L}_t}{\partial \hat{\vy}_t}$
    \State $\delta_{\mW} \gets \left(\frac{\partial \vh_t}{\partial \mW}\right)^\top \delta_{\vh}, \quad \mW \in \boldsymbol\theta$ \Comment{Gradient for \( \boldsymbol\theta \)}
    \State $\mW \gets \mW - \alpha \, \delta_{\mW},
    \quad \mW \in \boldsymbol\theta$     \Comment{Parameter update for \( \boldsymbol\theta \)}
    \State $\mV \gets \mV - \alpha \, \delta_{\mV}, \quad \mV \in \boldsymbol\phi$  \Comment{Parameter update for \( \boldsymbol\phi \)}
\EndFor
\end{algorithmic}
\end{algorithm}

\section{The RTRL Algorithm}
\label{appendix:rtrl-algorithm}
\begin{algorithm}[H]
\caption{The Real-time Recurrent Learning Algorithm}\label{alg:rtrl}
\begin{algorithmic}[1]
\State \textbf{Require}: Recurrent function \( f_{\boldsymbol\theta} \) with parameters \( \boldsymbol\theta \)
\State \textbf{Require}: Output function \( g_{\boldsymbol\phi} \) with parameters \( \boldsymbol\phi \)
\State \textbf{Require}: Data stream \( \mathcal{D} \), step size \( \alpha \)
\State \textbf{Initialize}: \( \boldsymbol\theta, \boldsymbol\phi \), hidden state \( \mathbf{h}_0 \gets \mathbf{0} \)
\State \textbf{Initialize}: Sensitivity matrix \( \mS^{\mW}_0 \gets \mathbf{0}, \forall \mW \in \boldsymbol\theta \) 
\Comment{ \( \mS^{\mW}_0 = \frac{\dd \mathbf{h}_0}{\dd \mW} \)}    
\For{\(t=1\dots\infty\)}
\State  $\delta_{\mV}\leftarrow \boldsymbol{0}, \forall \mV \in \boldsymbol\phi$, $\delta_{\mW}\leftarrow \boldsymbol{0}, \forall \mW \in \boldsymbol\theta$
    \State $\vx_t,\vy_t, \texttt{reset} \leftarrow \mathcal{D}$
    \If{\texttt{reset}}
    \State $\vh_{t-1} \leftarrow \boldsymbol{0}$
    \EndIf
    \State \( \mathbf{h}_t \gets f_\theta(\mathbf{h}_{t-1}, \mathbf{x}_t) \) 
    \Comment{Forward pass}
    
    \State Update sensitivity:
    $\mS^{\mW}_t \gets \frac{\partial \mathbf{h}_{t}}{\partial \mW} 
    + \frac{\partial \mathbf{h}_{t}}{\partial \mathbf{h}_{t-1}} \mS^{\mW}_{t-1}, \quad \forall \mW \in \boldsymbol\theta
    $

    \State \( \hat{\mathbf{y}}_t \gets g_{\boldsymbol\phi}(\mathbf{h}_t, \mathbf{x}_t) \)
    
    \State Compute \(\mathcal{L}_t = \mathcal{L}(\hat{\vy}_t, \vy_t)\)

    \State \( \delta_{\mV} \gets  \delta_{\mV} + \left(\frac{\partial \hat{\mathbf{y}}_t}{\partial \mV}\right)^\top\frac{\partial \mathcal{L}_t}{\partial \hat{\mathbf{y}}_t} , \quad \forall \mV \in \boldsymbol\phi \) \Comment{Gradient for \( \boldsymbol\phi \)}
    
    \State \( \delta_{\mW} \gets \delta_{\mW} + \left(\mS^{\mW}_t\right)^\top\left(\frac{\partial \hat{\mathbf{y}}_t}{\partial \mathbf{h}_t} \right)^\top\frac{\partial \mathcal{L}_t}{\partial \hat{\mathbf{y}}_t} 
    , \quad \forall \mW \in \boldsymbol\theta \) \Comment{Gradient for \( \boldsymbol\theta \)}
    \State \( \mW \gets \mW - \alpha\, \delta_{\mW} , \quad \forall \mW \in \boldsymbol\theta \) 
    \Comment{Parameter update for \( \boldsymbol\theta \)}
    \State \( \mV \gets \mV - \alpha\, \delta_{\mV}, \quad \forall \mV \in \boldsymbol\phi \) 
    \Comment{Parameter update for \( \boldsymbol\phi \)}
\EndFor
\end{algorithmic}
\end{algorithm}

% %%%%%%%%%%%%%%%%%%%%%%%%%%%%%%%%%%%%%%%%%%%%%%%%%%%%%%%

\section{The SRM Oscillatory Clock}
\label{appendix:srm-clock}

SRM uses a parameter-free oscillatory clock to make time available to its associative memory. Let $\vc_t\in\mathbb{R}^{2C}$ contain $C$ two-dimensional rotors with frequencies $\boldsymbol\omega=(\omega_1,\ldots,\omega_C)$. Each rotor follows the fixed rotation
\begin{align}
    \mR(\omega_i)
    &= \begin{bmatrix}
        \cos\omega_i & -\sin\omega_i \\
        \sin\omega_i & \phantom{-}\cos\omega_i
    \end{bmatrix}\in\mathbb{R}^{2\times2}, \\
    \vc_{t+1}^{(i)}
    &= \mR(\omega_i)\vc_t^{(i)},
    \qquad \vc_t^{(i)}\in\mathbb{R}^{2}.
\end{align}
The complete update is block diagonal,
\begin{align}
    \mR(\boldsymbol\omega)
    &= \diag\bigl(\mR(\omega_1),\ldots,\mR(\omega_C)\bigr)
    \in\mathbb{R}^{2C\times2C}, \\
    \vc_{t+1} &= \mR(\boldsymbol\omega)\vc_t.
    \label{eq:srm-clock}
\end{align}
At the beginning of a stream, every rotor starts at phase zero with norm $1/\sqrt{C}$, $\vc_0^{(i)}=(1,0)^\top/\sqrt{C}$. Consequently,
\begin{align}
    \vc_t^{(i)}
    = \frac{1}{\sqrt{C}}\begin{bmatrix}
        \cos(t\omega_i) \\
        \sin(t\omega_i)
    \end{bmatrix},
    \qquad \lVert\vc_t\rVert = 1 .
    \label{eq:srm-clock-phase}
\end{align}
The scale $1/\sqrt{C}$ matters because the clock and the content terms are added in the same address space: with unit rotors, the clock would enter the key at norm $\sqrt{C}$ while the content terms enter at $O(1)$, so time would dominate content in every write address. Normalizing the clock to unit norm puts the two terms on a comparable scale.
The clock is reset with the recurrent state at stream boundaries. It has no trainable parameters, is independent of the input, and preserves each rotor's norm. We choose periods geometrically from $2$ to a maximum period $P_{\max}$:
\begin{align}
    P_i
    &= 2^{\,1 + \frac{i}{C-1}\left(\log_2 P_{\max}-1\right)},
    &
    \omega_i &= \frac{2\pi}{P_i}.
    \label{eq:srm-clock-frequencies}
\end{align}
Short periods distinguish nearby steps, while longer periods provide a coarser signal over a longer horizon. The band is a resolution choice rather than a horizon choice, and it should not simply be made as large as possible: a rotor whose period greatly exceeds the horizon being resolved barely rotates, so it contributes an almost constant offset to every write address and drives distinct addresses toward collinearity, which degrades retrieval. Because the periods are geometrically spaced and mutually incommensurate, a short band already separates timesteps far beyond $P_{\max}$. We use $P_{\max}=64$ and $C=32$ rotors throughout. The clock enters both sides of associative retrieval. A write at step $s$ and a read at step $t$ use the addresses
\begin{align}
    \bar\vk_s &= \mK_x\vx_s + \mK_p\vx_{s-1} + \mK_c\vc_s,
    &
    \vq_t &= \mW_{q,x}\vx_t + \mW_{q,c}\vc_t,
    \label{eq:srm-clock-addresses}
\end{align}
with $\vk_s=\bar\vk_s/\lVert\bar\vk_s\rVert$. Here, $\mK_x$, $\mK_p$, and $\mK_c$ are fixed projections, whereas $\mW_{q,x}$ and $\mW_{q,c}$ are learned. The clock, therefore, supplies a time address from initialization; learning determines how the query combines that signal with the content terms when retrieving a stored value.

The rotation geometry also makes relative offsets available. For each rotor,
\begin{align}
    \left\langle \vc_t^{(i)},\vc_s^{(i)}\right\rangle
    = \tfrac{1}{C}\cos\bigl((t-s)\omega_i\bigr).
    \label{eq:srm-clock-relative-position}
\end{align}
The distinction is clearest for a single rotor, for which we drop the scale $1/\sqrt{C}$ to reduce clutter. A fixed query using the initial clock phase $\vc_0=(1,0)^\top$ gives
\begin{align}
    \vc_0^\top\vc_s = \cos(s\omega),
\end{align}
which identifies the write by its absolute position $s$ relative to the stream reset. In contrast, a query using the current clock phase gives
\begin{align}
    \vc_t^\top\vc_s = \cos\bigl((t-s)\omega\bigr),
\end{align}
which identifies the write by its separation $t-s$ from the current read. Thus, $\vc_0$ is an absolute time reference, whereas the moving $\vc_t$ is a relative time reference. The learned query maps can use either comparison or combine them with content.

%%%%%%%%%%%%%%%%%%%%%%%%%%%%%%%%%%%%%%%%%%%%%%%%%%%%%%%%
\section{Experimental Details}
\label{appendix:exp-details}

\subsection{Baselines}
\label{appendix:baselines}
The task suite (Figures~\ref{fig:suite_init} and~\ref{fig:suite_train}) compares SRM with GLA \citep{yang2024gla}, KDA \citep{team2025kimi}, and RTU \citep{elelimy2024realtime}. The associative T-Maze compares SRM with RTU, and the language-modeling experiment compares SRM with GLA, KDA, RTU, and eLSTM
\citep{irie2024exploring}.

The original designs of GLA and KDA use a low-rank parameterization of some weights. In our experiments, we parameterize all weights with full-rank to isolate away any effects and implications of the low-rank parameterization choice, as SRM uses full-rank parameterization everywhere. For KDA, we use a convolution kernel size of two rather than the original four, matching that of SRM. We initialize every units (including SRM, as discussed in Section~\ref{section:srm-transition}) with the memory-preserving initialization strategy for the input-dependent gates---weights initialized to 0, and bias initialized to the LRU ring distribution. 

In every experiment, we match the size of the baseline units to SRM by FLOPs per timestep. Nonlinear operations are included in this computation, using a consistent FLOPs convention.

\subsection{Diagnostics}
\label{appendix:diagnostics}

Throughout the experiments in section \ref{sec:diagnostics}, we keep the architecture minimal to minimize confounding expressivity augmented by non-recurrent layers. This design allows us to isolate the capabilities that are provided by the recurrent unit itself. We use a frozen input projection (initialized orthogonally) of width 32, followed by a layer of the investigated recurrent unit, followed by a parameter-free RMSNorm layer (necessary for stable gradient flow), followed by a two-layer MLP classifier with a hidden dimension of 32 and GELU activation.

Table~\ref{tab:task-suite-capacity} specifies the state size of SRM and baseline units used in this architecture. FLOPs matching includes the cost of the recurrent unit's forward pass and one-step truncated-BPTT backward pass, but excludes the cost of external input projection, classifier, optimizer updates, and RTRL trace maintenance.  BPTT and RTRL variants use identical model size. 
\begin{table}[ht]
    \centering
    \caption{Size of the recurrent unit of SRM and baselines used throughout the diagnostics experiments. SRM uses 8 rotors for the clock and 8 tracking dimensions of the tracking channel. }
    \label{tab:task-suite-capacity}
    \begin{tabular}{lrrrr}
        \hline
        Unit & Input width & State width & Read heads & tracking dimension \\
        \hline
        SRM & 32 & 32 & 2 & 8 \\
        GLA    & 32 & 42 & 2 & -- \\
        KDA    & 32 & 20 & 2 & -- \\
        RTU    & 32 & 60 & -- & -- \\
        eLSTM  & 32 & 45 & -- & -- \\
        \hline
    \end{tabular}
\end{table}

\subsection{Task Suite}
\label{appendix:task-suite}

We probe the memory capabilities of each recurrent unit using a suite of ten synthetic tasks. Each task specifies a presentation alphabet $\Sigma$, a distribution over presentation sequences, and a target to predict. The presentation length $L$ is sampled uniformly from $\{5,6,7,8\}$. An example has the form
\[
    x_1,\ldots,x_L,\bullet,q_1,\ldots,q_m,
\]
where $\bullet\notin\Sigma$ marks the end of the presentation and the $q_i$ are optional task queries. Supervision occurs at $\bullet$ when $m=0$ and at the final query otherwise.

The tasks are defined as follows:

\begin{description}
    \item[\texttt{Presence}]
    The presentation contains distinct symbols. Given $q\in\Sigma$, predict whether $q$ appeared.

    \item[\texttt{Successor}]
    The presentation contains distinct symbols. Given a presented symbol $q$ other than the final symbol, predict the symbol immediately following it.

    \item[\texttt{Predecessor}]
    The presentation contains distinct symbols. Given a presented symbol $q$ other than the initial symbol, predict the symbol immediately preceding it.

    \item[\texttt{Count}]
    Given $q\in\Sigma$, predict how many times $q$ appeared. The target count is sampled from $\{0,1,2,3\}$, and no symbol appears more than three times.

    \item[\texttt{Order}]
    The presentation contains distinct symbols. Given two presented symbols $q_1$ and $q_2$, predict whether $q_1$ appeared before $q_2$.

    \item[\texttt{Position}]
    The presentation contains distinct symbols from $\Sigma=\{0,\ldots,7\}$. Let $p_i$ refer to the position $i < L$ from the presentation. Given $p_i$, predict the symbol presented at position $i$.

    \item[\texttt{Parity}]
    Symbols are sampled independently from $\{0,1\}$. Predict $\sum_{i=1}^{L}x_i \bmod 2$.

    \item[\texttt{Orient}]
    Let $\mathrm{rT}$ denote a quarter (clockwise) rotation and $\mathrm{rF}$ a reflection (along the axis perpendicular to the heading). Beginning at the heading $h_0$, apply the presented rotations and reflections and predict the final heading in $\{h_0,h_1,h_2,h_3\}$.
    
    \item[\texttt{Shell}]
    Let $s_{01}$, $s_{02}$, and $s_{12}$ denote swaps of the corresponding pairs of cups. Beginning with the object under cup $c_0$, apply the presented swaps and predict its final cup in $\{c_0,c_1,c_2\}$.

    \item[\texttt{Mod3}]
    Symbols are sampled independently from $\{0,1,2\}$. Predict $\sum_{i=1}^{L}x_i \bmod 3$.
\end{description}

\begin{table*}[ht]
\centering
\scriptsize
\renewcommand{\arraystretch}{1.25}
\begin{tabular}{@{}lllll@{}}
\toprule
\textbf{Task}
& $\boldsymbol{|\Sigma|}$
& \textbf{Chance}
& \multicolumn{2}{c}{\textbf{Examples}} \\
\cmidrule(lr){4-5}
&&& \textbf{Sequence} & \textbf{Target} \\
\midrule

\texttt{Presence}
& 16 & $0.594$
& \begin{tabular}[c]{@{}l@{}}
  $\langle15,9,4,0,7,\bullet,8\rangle$\\
  $\langle10,4,8,5,15,\bullet,4\rangle$
  \end{tabular}
& \begin{tabular}[c]{@{}c@{}}$0$\\$1$\end{tabular} \\

\addlinespace
\texttt{Successor}
& 8 & $0.190$
& \begin{tabular}[c]{@{}l@{}}
  $\langle7,1,2,6,0,\bullet,1\rangle$\\
  $\langle5,1,2,7,6,3,4,0,\bullet,2\rangle$
  \end{tabular}
& \begin{tabular}[c]{@{}c@{}}$2$\\$7$\end{tabular} \\

\addlinespace
\texttt{Predecessor}
& 8 & $0.190$
& \begin{tabular}[c]{@{}l@{}}
  $\langle7,1,2,6,0,\bullet,2\rangle$\\
  $\langle5,1,2,7,6,3,4,0,\bullet,7\rangle$
  \end{tabular}
& \begin{tabular}[c]{@{}c@{}}$1$\\$2$\end{tabular} \\

\addlinespace
\texttt{Count}
& 4 & $0.250$
& \begin{tabular}[c]{@{}l@{}}
  $\langle0,1,0,1,3,\bullet,3\rangle$\\
  $\langle3,0,2,1,0,1,0,\bullet,0\rangle$
  \end{tabular}
& \begin{tabular}[c]{@{}c@{}}$1$\\$3$\end{tabular} \\

\addlinespace
\texttt{Order}
& 8 & $0.500$
& \begin{tabular}[c]{@{}l@{}}
  $\langle7,1,2,6,0,\bullet,2,0\rangle$\\
  $\langle7,3,4,2,1,\bullet,3,1\rangle$
  \end{tabular}
& \begin{tabular}[c]{@{}c@{}}$1$\\$1$\end{tabular} \\

\addlinespace
\texttt{Position}
& 8 & $0.159$
& \begin{tabular}[c]{@{}l@{}}
  $\langle7,1,2,6,0,\bullet,p_2\rangle$\\
  $\langle5,1,2,7,6,3,4,0,\bullet,p_4\rangle$
  \end{tabular}
& \begin{tabular}[c]{@{}c@{}}$2$\\$6$\end{tabular} \\

\addlinespace
\texttt{Parity}
& 2 & $0.500$
& \begin{tabular}[c]{@{}l@{}}
  $\langle1,1,0,1,1,\bullet\rangle$\\
  $\langle1,1,1,1,0,0,1,0,\bullet\rangle$
  \end{tabular}
& \begin{tabular}[c]{@{}c@{}}$0$\\$1$\end{tabular} \\

\addlinespace
\texttt{Orient}
& 2 & $0.250$
& \begin{tabular}[c]{@{}l@{}}
  $\langle\mathrm{rF},\mathrm{rF},\mathrm{rT},\mathrm{rF},
  \mathrm{rF},\bullet\rangle$\\
  $\langle\mathrm{rT},\mathrm{rF},\mathrm{rT},\mathrm{rT},
  \mathrm{rF},\mathrm{rT},\bullet\rangle$
  \end{tabular}
& \begin{tabular}[c]{@{}c@{}}$h_1$\\$h_0$\end{tabular} \\

\addlinespace
\texttt{Shell}
& 3 & $0.333$
& \begin{tabular}[c]{@{}l@{}}
  $\langle s_{01},s_{12},s_{01},s_{02},s_{01},\bullet\rangle$\\
  $\langle s_{01},s_{01},s_{12},s_{01},s_{12},s_{01},\bullet\rangle$
  \end{tabular}
& \begin{tabular}[c]{@{}c@{}}$c_1$\\$c_2$\end{tabular} \\

\addlinespace
\texttt{Mod3}
& 3 & $0.333$
& \begin{tabular}[c]{@{}l@{}}
  $\langle0,2,0,1,0,\bullet\rangle$\\
  $\langle1,1,0,2,2,0,0,1,\bullet\rangle$
  \end{tabular}
& \begin{tabular}[c]{@{}c@{}}$0$\\$1$\end{tabular} \\

\bottomrule
\end{tabular}
\caption{
The synthetic task suite. $|\Sigma|$ denotes the presentation alphabet size. Chance is the expected accuracy of a task-appropriate guesser, with variable presentation length accounted for.
}
\label{tab:task-suite1}
\end{table*}

\subsection{Memory Capabilities without Learning}
\label{appendix:init-exp}
For each task, we construct a fixed bank of 480 distinct examples, with 120 examples at each presentation length $L\in\{5,6,7,8\}$. The examples are generated according to the task distributions described in Appendix~\ref{appendix:task-suite}. All units and initialization seeds are evaluated on the same bank for each task. For \texttt{Parity} and \texttt{Orient}, the 480 examples exhaust the complete presentation space.

To measure retention, we insert $n$ repetitions of a waiting symbol $\circ\notin\Sigma$ between the presentation and the end-of-presentation symbol. Thus, an example at delay $n$ has the form
\[
    x_1,\ldots,x_L,
    \underbrace{\circ,\ldots,\circ}_{n\text{ waiting steps}}, 
    \bullet, 
    \underbrace{q_1,\ldots,q_m}_{\text{task queries}}.
\]

We supply a dataset of 480 examples, one for each of $n\in\{0,100,200,500\}$. The recurrent state is reset before every example, and the same set of 480 examples unfolds the recurrent state. All recurrent roll-outs are performed in double precision. After the final symbol, the recurrent state is flattened into a probe feature vector. We denote its size to be $d$. Stacking these vectors produces, for each delay, a feature matrix
\[ X_n\in\mathbb{R}^{480\times d}.\]

For \texttt{Parity} and \texttt{Orient}, the 480 presentations comprise the complete population for lengths $L\in\{5,6,7,8\}$. We therefore fit and evaluate the ridge probe on the complete population using the fixed regularization parameter $\kappa=10^{-4}$.
Because the number of presentations differs across lengths, we weight them accordingly to ensure each length contributes equally to the final accuracy. The final score measures linear decodability over the complete finite population, rather than out-of-sample generalization.

For tasks other than \texttt{Parity} and \texttt{Orient}, to ensure out-of-sample generalization, we evaluate linear ridge probes using length-stratified nested four-fold cross-validation. 
In detail, both the outer and inner folds are stratified by presentation length. For each outer fold, we hold out $25\%$ of the examples for testing. We then perform four-fold inner cross-validation with the remaining $75\%$, for each candidate value of the ridge regularization parameter
\[
    \kappa\in
    \left\{
    10^{-8},10^{-7},10^{-6},10^{-5},10^{-4},10^{-3},10^{-2},10^{-1},
    10^0,10^1,10^2,10^3,10^4
    \right\}.
\]
We select the $\kappa$ value with the highest mean score across the four inner validation folds.

With the selected $\kappa$, the probe is refitted on the complete outer training set and evaluated on the held-out outer test fold. 
Features are standardized using statistics computed only from this corresponding training set; the intercept is not regularized. Each example receives exactly one out-of-fold prediction. For each outer validation fold, the accuracy is computed separately at each presentation length and then averaged equally across the four lengths. We report the mean accuracy across the four outer validation folds.
 
We repeat the experiment for 10 different instantiations of each recurrent unit. We use model capacity as specified in Table~\ref{tab:task-suite-capacity}. We instantiate every unit using each specific initialization scheme designed for long retention. That includes the ring initialization, with $r_\text{min}=0.97$ and $r_\text{max}=0.9995$. At the same time, we control for the expressivity of the unit by avoiding initializing weights to 0 (weights at 0 would potentially sterilize some capabilities), ensuring the experiment would accurately find a unit's nascent capabilities.

\subsection{Reliability of one-step BPTT in learning SRM parameters}
\label{appendix:gradient-exp}

We introduce the alignment score $\rho$ in Section~\ref{subsec:gradients} as a measurement of one-step BPTT gradient reliability, relative to Full BPTT. We explain it more here. Under full BPTT, when a loss occurs at timestep $T$, the error is propagated backward to produce a distinct gradient contribution at each timestep, \(g_t\) for each learnable parameter. Since the parameter is reused across time, it receives the same update based on accumulated gradients. Full BPTT computes the gradients for a particular parameter by taking the aggregate of the gradients accumulated over time:
\(g_\mathrm{full} = \sum_{t=0}^{T} g_t.\)

One-step BPTT retains only the gradients contribution at the most recent step \(g_\mathrm{one\_step} = g_T\), thereby omitting all the gradient contributions of earlier steps, whose sum we denote by
\[g_\mathrm{earlier\_steps} = \sum_{t=0}^{T-1} g_t = g_\mathrm{full} - g_\mathrm{one\_step}.\]
While an optimizer can scale the size of a gradient update, it cannot recover the \textit{direction} of the gradients that the gradient estimator omits. Thus, learning any dependency longer than one step is only reliable if the gradient estimator can reliably produce gradients that are aligned with \(g_\mathrm{earlier\_steps}\) directionally. We measure this alignment, for a particular parameter, with the Frobenius cosine similarity, and denote the score as $\rho$:
\[
\rho_\mathrm{param} = \frac{
\left\langle g_{\mathrm{one\_step}},g_{\mathrm{earlier\_steps}}\right\rangle_F
}{
\left\lVert g_{\mathrm{one\_step}}\right\rVert_F
\left\lVert g_{\mathrm{earlier\_steps}}\right\rVert_F
},
\]
where \(\langle A,B\rangle_F\) denotes the Frobenius inner product and $||A||_F$ denotes the Frobenius norm. A positive value is a promising sign that one-step BPTT may perform credit assignment for this parameter beyond one timestep. A negative value indicates that this gradient will move the parameter against learning longer-than-one-step memory. Frequent negative alignment would suggest that learning with one-step BPTT is harmful and unreliable.

\begin{figure}[ht]
    \centering
    \includegraphics[width=0.9\linewidth]{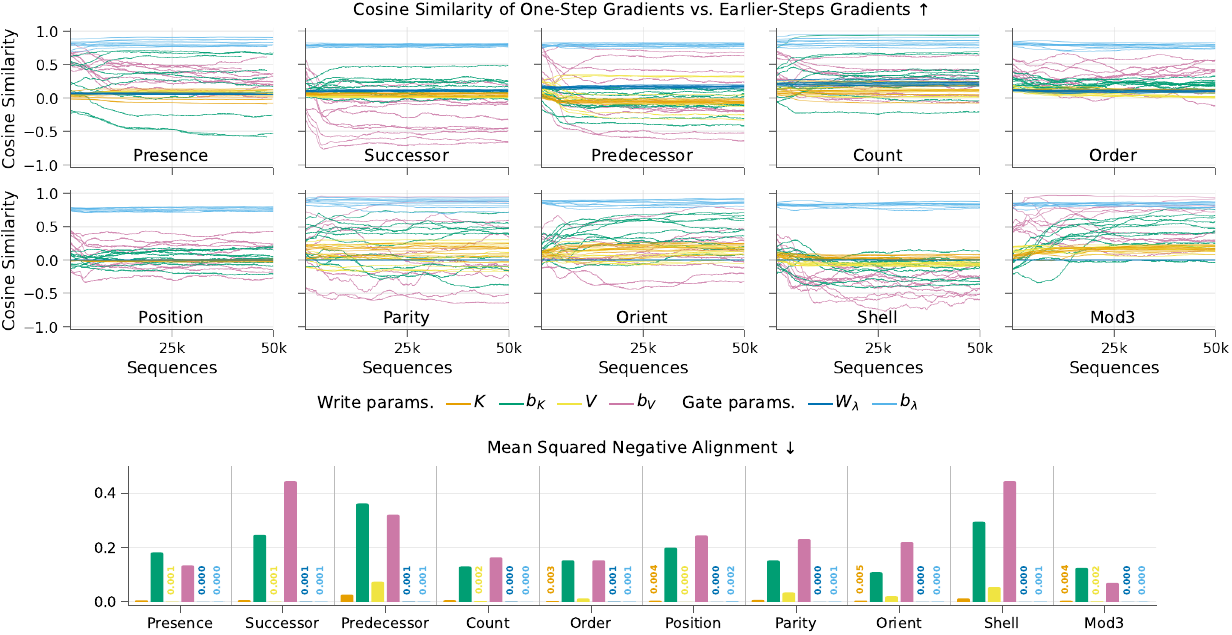}
    \caption{(Top) Alignment score for parameters of SRM over the training horizon of the readout, across the task suite, with all of non-readout parameters fixed throughout this training. Results are smoothed with a 5000-point centered rolling average. (Bottom) Squared negative alignment, averaged over training steps and runs. This value ranges from 0 to 1, with higher values indicating that local updates oppose the gradient contributions from earlier steps.}
    \label{fig:alignment_full_suite}
\end{figure}

We examine how $\rho$ behaves with respect to the task structure and the condition of the query and readout, in Experiment~\ref{subsec:gradients}, by training the readout and query parameters while keeping everything else frozen. Given an initialization of SRM, this protocol allows the loss to evolve in a controlled manner, while keeping the score comparable throughout training. The alignment score $\rho$ is obtained for selected parameters of SRM by autograd and cosine similarity computation. We report the score $\rho$ for selected parameters throughout this training procedure for selected tasks in Figure~\ref{fig:alignment_summary}, and for the whole suite, here in Figure~\ref{fig:alignment_full_suite}, for 10 initializations of SRM.

We note that this measurement $\rho$ is more meaningful for our purpose than the cosine similarity between $g_{\mathrm{one\_step}}$ and $g_\mathrm{full}$. The reason is that $g_\mathrm{full}$ contains the contribution of the one-step gradients, in unknown amount, inflating the directional alignment between them. This inflated score would be misleading for analyzing how well one-step BPTT can perform credit assignment beyond one step.

The squared negative alignment is obtained with $\min(0, \rho)^2$. It penalizes negative alignment quadratically while treating zero or positive alignment as neutral. This metric is deliberately risk-sensitive, emphasizing severe negative alignment that could lead to detrimental optimization updates, while remaining tolerant of small deviations below zero. We obtain the value of the mean squared negative alignment by taking the average across all training steps and runs.

\subsection{Memory Capabilities with Streaming Learning}
\label{appendix:suite-exp}
In this experiment, we use the ring initialization for all units, with $r_\text{min}=0.9$ and $r_\text{max}=0.999$. We use the StreamingOptimizer \citep{elsayed2026streaming} with $\beta_2 = 0.99$. We perform an iterative search for hyperparameters of StreamingOptimizer in this experiment. We search over $\beta_2 \in \{0.0, 0.9, 0.999, 0.9995\}$, at the learning rate of $1\times10^{-3}$, and find that $\beta_2 = 0.99$ is the best across the task suite and recurrent units.

For each combination of units and tasks, we sweep over 3 seeds to choose the better learning rate between $1\times10^{-3}$ and $3\times10^{-3}$. With the chosen learning rate, we report results over 10 different seeds. This process is done separately for BPTT and RTRL.

For each training run, we aggregate per-sequence accuracy using non-overlapping bins of 100 sequences. We then average the binned accuracies across 10 independent runs and apply a 50-point moving average to obtain the learning curves in Figure~\ref{fig:suite_train}.

\subsection{Streaming Learning on MQAR}
\label{appendix:mqar}
\begin{figure}[ht]
    \centering
    \includegraphics[width=0.88\linewidth]{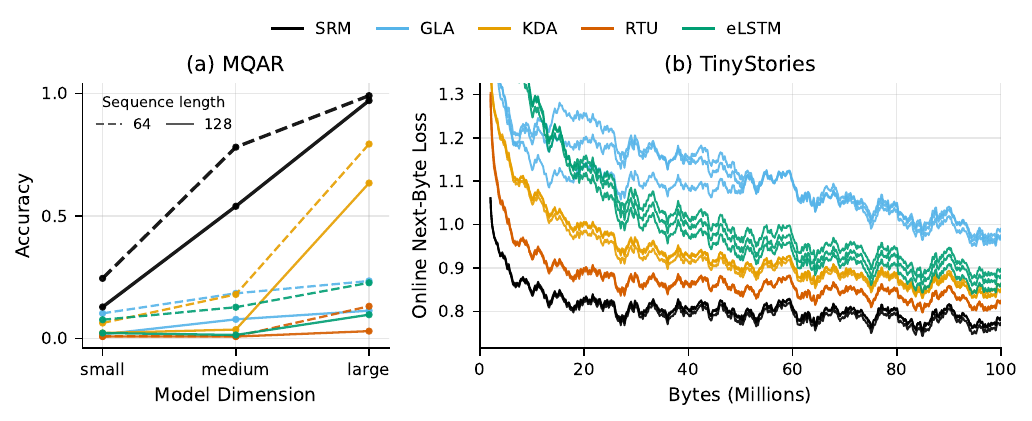}
    \caption{(a) MQAR: Average accuracy across three model dimensions, with dashed and solid lines denoting the shorter and longer task settings. (b) Online next-byte loss on TinyStories, over $100$M training bytes. Each line is one of three independent runs. All models use networks that have around $13$M parameters.}
    \label{fig:tasks}
\end{figure}

In our experiment, we reduce the alphabet size of standard MQAR configuration \citep{arora2024zoology}  from $8,192$ to $256$. This is necessary to allow any streaming learner we test to produce meaningful results. We also reserve a symbol, instead of using random non-query symbols, for waiting timesteps. The two settings of MQAR we report in Table~\ref{tab:tasks} and Figure~\ref{fig:tasks}(a)---64 and 128 sequence length---use 4 and 8 training pairs respectively, matching the standard setup. We use an architecture of $2$ blocks (Figure~\ref{fig:arch}), followed by a final RMSNorm  and a non-linear, GeLU activated classifier readout.

We set the model dimension of the baseline units to approximately match three SRM capacity levels in FLOPs per timestep, as listed in Table~\ref{tab:mqar-capacity}. The model dimension values specify the width of all of the following layers: input projection, recurrent state, and the hidden layer of the classifier readout.

For SRM, the number of clock rotors and the dimensions of the tracking channel are both scaled to $\frac{1}{4}$ the model dimension.  Matching FLOPS includes counting all model layers in one forward pass, one-step truncated-BPTT backward pass, and one optimizer update, including estimated nonlinear-operation costs. The same configurations are used for both variants of the task.

\begin{table}[ht]
    \scriptsize
    \centering
    \caption{FLOPs-matched model dimensions for SRM and baselines, for MQAR.}
    \label{tab:mqar-capacity}
    \begin{tabular}{lrrr}
        \hline
        & \multicolumn{3}{c}{Model dimension} \\
        Model & Small & Medium & Large \\
        \hline
        SRM   & 32 & 64 & 128 \\
        GLA   & 40 & 76 & 156 \\
        KDA   & 36 & 72 & 144 \\
        RTU   & 42 & 90 & 184 \\
        eLSTM & 37 & 77 & 157 \\
        \hline
    \end{tabular}
\end{table}

In MQAR, a loss occurs at every query timestep, which can appear multiple times within a training sequence. We accumulate gradients at each query timestep and update parameters once at the end of the sequence, keeping parameters fixed within each sequence.

We use the StreamingOptimizer with $\beta_2 = 0.999$. This value is chosen from a grid search over $\beta_2 \in \{0.9, 0.999, 0.9995\}$, at the learning rate of $1\times10^{-3}$, on the MQAR variant with 64 sequence length, for all units at the specified small model-dimension variant. With $\beta_2 = 0.999$, we sweep over 3 runs to find the best learning rate in $\{3\times10^{-4}, \, 1\times10^{-3}, \, 3\times10^{-3}\}$, for each combination of unit and size, on the MQAR variant of 64 length. Then, we train each unit over $50,000$ sequences at each task setting, and report the results in Table~\ref{tab:tasks} and Figure~\ref{fig:tasks}(a) as the average accuracy of the final 10\% training sequences, averaged over 10 independent runs.

Figure~\ref{fig:tasks} plots the full results summarized in Table~\ref{tab:tasks}: the MQAR accuracies across model dimensions and the complete TinyStories online-loss curves of Section~\ref{subsec:language}.

\subsection{Associative T-Maze}
\label{appendix:assoc-tmaze}

The observation is a $12$-dimensional one-hot vector with $8$ key slots, $2$ value slots (\texttt{left}, \texttt{right}), a blank slot, and a query flag; at the junction, the queried key's slot and the query flag are both set. Each episode draws $K$ distinct keys uniformly from the $8$ symbols, an independent uniform value for each key, and one of the $K$ pairs uniformly as the query, so an episode lasts $2K+D+1$ steps with $D=10$. There are two actions. Actions before the junction are ignored, and the junction action ends the episode with reward $+1$ if it matches the queried value and $-1$ otherwise; all other rewards are $0$.

The policy and value functions use separate networks with identical structure: a linear input projection with bias, one streaming block (Figure~\ref{fig:arch}) with model width $M=64$ and state width $S=64$, a final RMSNorm, and a linear head. SRM's tracking strength $(\vw_\beta,b_\beta)$ is trained with one-step BPTT, whose sensitivity is a $G\times(d+1)$ matrix and cheap for $G=16$; the query maps and the retention receive the one-step gradient. RTU is also trained with one-step BPTT. Both units use the ring initialization with $r_\text{min}=0.97$ and $r_\text{max}=0.9995$. The memoryless baseline is the SRM agent with its recurrent state reset before every step. The final online success rates of SRM are $0.998$, $0.996$, and $0.997$ for $K=2$, $4$, and $8$; the learning curves in Figure~\ref{fig:assoc-tmaze} are averaged over blocks of $1{,}000$ episodes.

Every step produces one update of each network from the TD error $\delta_t=r_{t+1}+\gamma\hat v(s_{t+1})-\hat v(s_t)$, with $\hat v(s_{t+1})=0$ at the junction. Each parameter keeps an accumulating eligibility trace $\ve\leftarrow\gamma\lambda_\text{tr}\ve+\nabla$, reset at the end of each episode, and moves by $\alpha\,\delta_t\ve/(\mu+\epsilon)$, where $\mu$ is a running maximum of $|\delta_t\ve|$ with decay $0.99995$ and $\epsilon=10^{-8}$. The actor's gradient is that of $\log\pi(a_t\mid s_t)$ plus the entropy bonus multiplied by $\operatorname{sign}(\delta_t)$. Table~\ref{tab:assoc-tmaze-hparams} lists the settings.

For each unit and each $K$, we run the four combinations of step-size pair and entropy coefficient in Table~\ref{tab:assoc-tmaze-hparams} with three seeds and choose the combination with the highest mean cumulative online return. We then run the chosen combination on ten new seeds, which are the ones reported in Figure~\ref{fig:assoc-tmaze}. The memoryless baseline uses SRM's chosen combination. Because each episode's return is $\pm1$, the online success rate over any block of episodes is exactly $\tfrac12(1+\bar R)$, where $\bar R$ is the mean return over the block.

\begin{table}[ht]
\centering
\scriptsize
\begin{tabular}{ll}
\toprule
\textbf{Setting} & \textbf{Value} \\
\midrule
Discount $\gamma$, trace decay $\lambda_\text{tr}$ & $0.99$, $0.8$ \\
Step sizes (critic head $\alpha$, recurrent networks $\alpha_\text{rec}$) & $(10^{-2}, 10^{-3})$ or $(10^{-3}, 10^{-2})$ \\
Actor head step size & $\alpha/2$ \\
Entropy coefficient & $0.1$ \\
Training episodes & $20{,}000$ ($30{,}000$ for $K=8$) \\
\bottomrule
\end{tabular}
\caption{Associative T-Maze settings. The step sizes and the entropy coefficient are selected per unit and per $K$.}
\label{tab:assoc-tmaze-hparams}
\end{table}

\subsection{Streaming Language Modeling}
\label{appendix:lm-exp}

\begin{figure}[ht]
    \centering
    \begin{subfigure}{0.20\textwidth}
        \centering
        \includegraphics[width=\linewidth]{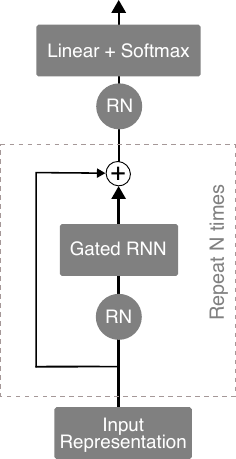}
        \caption{Architecture}
    \end{subfigure}
    \hspace{1cm}
    \begin{subfigure}{0.33\textwidth}
        \centering
        \includegraphics[width=\linewidth]{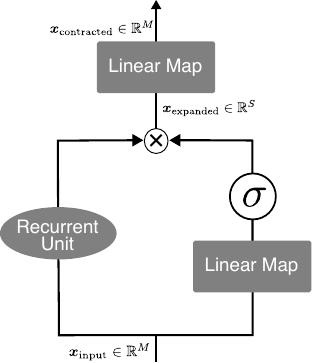}
        \caption{Gated RNN}
    \end{subfigure}
    \caption{Architecture used throughout the paper. RN denotes RMSNorm. $M$ is the model dimension and $S$ is the state dimension. $\sigma$ denotes a sigmoid nonlinearity.}
    \label{fig:arch}
\end{figure}

The stream is the TinyStories training set \citep{eldan2023tinystories} read as raw bytes, so the vocabulary has $256$ symbols. Stories are presented in dataset order, and the recurrent state is reset at each story boundary. Each model consists of a byte embedding, four blocks (Figure~\ref{fig:arch}), and a linear output head. Every byte produces one StreamingOptimizer update with $\beta_1=0$, $\beta_2=0.99995$, $\epsilon=10^{-8}$, and step size $3\times10^{-4}$, from the gradient of the next-byte cross-entropy computed with one-step BPTT. The online loss is logged as the mean over consecutive windows of $1{,}000$ bytes, each byte being scored before the update it produces. Each unit is run with three seeds for $100$M bytes. Table~\ref{tab:lm-widths} lists the widths.

\begin{table}[ht]
\centering
\scriptsize
\begin{tabular}{lcc}
\toprule
\textbf{Unit} & \textbf{Model width $M$} & \textbf{State width $S$} \\
\midrule
RTU & $1024$ & $1024$ \\
eLSTM & $1024$ & $512$ \\
GLA & $768$ & $512$ \\
KDA & $768$ & $512$ \\
SRM & $768$ & $512$ \\
\bottomrule
\end{tabular}
\caption{Widths of the four-layer streaming language models. All models have approximately the same number of parameters around $13$M.}
\label{tab:lm-widths}
\end{table}

In every streaming block, the gate and the output projection have Xavier-uniform weights, and the RMSNorm scales start at one. Recurrent units other than SRM use the initialization of their reference implementations. All units are trained with one-step BPTT: the recurrent state is carried from byte to byte, but the gradient of each update stops at the previous state.

SRM's recurrent parameters are the exception to the above: they follow the scheme given in Section~\ref{section:srm}. Specifically, $\mK_x$, $\mK_p$, $\mK_c$, $\mV_x$, $\mV_p$, and $\mU$ are semi-orthogonal; each head's $\mW_{q,x}$ is semi-orthogonal scaled by $1/\sqrt{d}$ while $\mW_{q,c}$ is zero; $\mW_\lambda=\mathbf{0}$ with $b_{\lambda,s}=\log(-\log r_s)$ from the ring draw; and the tracking strength starts at $\beta_t=1.9$ ($\vw_\beta=\mathbf{0}$, $b_\beta=2\operatorname{arctanh}(0.9)$) and is learned. We use $r_\text{min}=0.0$, $r_\text{max}=1.0$ in the streaming language modeling experiments. The clock uses $P_{\max}=64$ with $C=32$ rotors, the tracking channel has width $G=16$, and SRM uses $N_\text{h}=4$ read heads with value width $d_v=d/N_\text{h}$ so that the read has the same width as the block's residual stream. In Figure~\ref{fig:arch}, SRM's input width $d$ is the model dimension $M$, and its state dimension $S$ is the number of memory rows.

%%%%%%%%%%%%%%%%%%%%%%%%%%%%%%%%%%%%%%%%%%%%%%%%%%%%%%%%

\subsubsection{Optimizer Choice}
\label{appendix:lm-optimizer}

\begin{figure}[ht]
    \centering
    \includegraphics[width=0.5\linewidth]{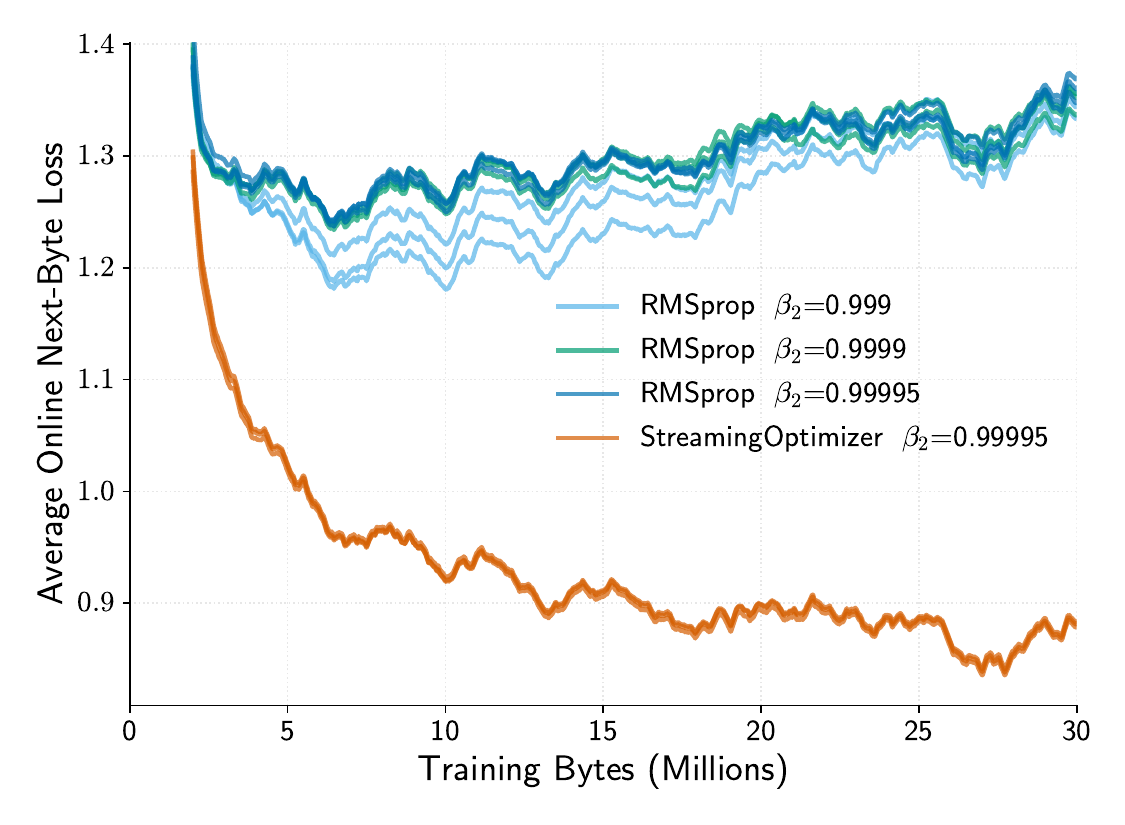}
    \caption{Online next-byte loss of RTU trained with one-step BPTT under RMSprop with different $\beta_2$ and under StreamingOptimizer (equivalent to Adamax with $\beta_1=0$), over the first $30$M bytes.}
    \label{fig:lm-optimizer}
\end{figure}

Figure~\ref{fig:lm-optimizer} compares RMSprop \citep{tieleman2012lecture} across second-moment decay rates $\beta_2$ with StreamingOptimizer \citep{elsayed2026streaming}, which is equivalent to Adamax \citep{kingma2014adam} with $\beta_1=0$, for RTU trained with one-step BPTT on $30$M bytes. With RMSprop, the online loss falls to about $1.2$ within the first few million bytes and then rises to about $1.35$ by $30$M bytes for every $\beta_2$, whereas with StreamingOptimizer it keeps decreasing and ends near $0.88$.

\subsubsection{Architectures of the Scaling Sweep}
\label{appendix:lm-scaling}

\begin{figure}[ht]
    \centering
    \includegraphics[width=0.85\linewidth]{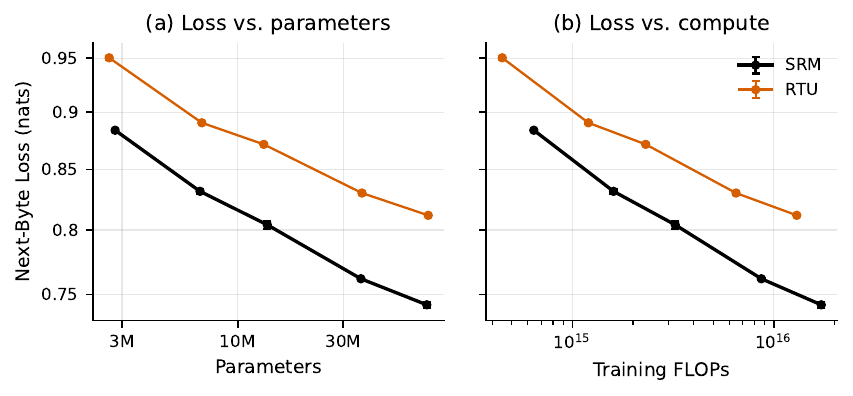}
    \caption{Next-byte loss of SRM and RTU across the scaling sweep, against (a) the number of parameters and (b) the training FLOPs. Error bars show the standard error of the mean. The configurations are listed in Tables~\ref{tab:lm-scaling-srm} and~\ref{tab:lm-scaling-rtu}.}
    \label{fig:lm-scaling}
\end{figure}

Tables~\ref{tab:lm-scaling-srm} and~\ref{tab:lm-scaling-rtu} list the SRM and RTU configurations of the scaling sweep. The configurations of the two units are matched in parameter count, so the per-layer recurrent state they carry differs. Figure~\ref{fig:lm-scaling} shows the resulting next-byte loss: SRM reaches a lower loss than RTU at every size, both at matched parameters and at matched training FLOPs.

\begin{table}[ht]
\scriptsize
\centering
\begin{tabular}{ccccccc}
\toprule
\textbf{Params} & \textbf{Layers} & \textbf{Width $M$} & \textbf{Rows $S$} & \textbf{Memory $S\times M/4$} & \textbf{State / layer} & \textbf{Total state} \\
\midrule
$2.8$M & $3$ & $384$ & $256$ & $256\times96$ & $25{,}040$ & $75$K \\
$6.8$M & $4$ & $512$ & $384$ & $384\times128$ & $49{,}744$ & $199$K \\
$13.6$M & $4$ & $768$ & $512$ & $512\times192$ & $99{,}152$ & $397$K \\
$36.0$M & $6$ & $1024$ & $704$ & $704\times256$ & $181{,}328$ & $1.09$M \\
$71.5$M & $8$ & $1280$ & $832$ & $832\times320$ & $267{,}600$ & $2.14$M \\
\bottomrule
\end{tabular}
\caption{SRM configurations of the scaling sweep. $M$ is the model width and $S$ is the number of memory rows.}
\label{tab:lm-scaling-srm}
\end{table}

\begin{table}[ht]
\scriptsize
\centering
\begin{tabular}{cccccc}
\toprule
\textbf{Params} & \textbf{Layers} & \textbf{Width $M$} & \textbf{State width $S$} & \textbf{State / layer} & \textbf{Total state} \\
\midrule
$2.6$M & $3$ & $512$ & $512$ & $512$ & $1.5$K \\
$6.9$M & $4$ & $736$ & $736$ & $736$ & $2.9$K \\
$13.1$M & $4$ & $1024$ & $1024$ & $1{,}024$ & $4.1$K \\
$36.4$M & $6$ & $1408$ & $1408$ & $1{,}408$ & $8.4$K \\
$72.6$M & $8$ & $1728$ & $1728$ & $1{,}728$ & $13.8$K \\
\bottomrule
\end{tabular}
\caption{RTU configurations of the scaling sweep. The state of width $S$ is held as $S/2$ complex units.}
\label{tab:lm-scaling-rtu}
\end{table}

\subsubsection{Sample Story from Streaming SRM}
\label{appendix:lm-sample}

The following story was generated byte by byte by the SRM language model of Section~\ref{subsec:language} after $100$M training bytes, from the prompt ``Once upon a time'' with temperature $0.7$ and top-$k$ sampling with $k=50$. The text is reproduced verbatim, including its errors.

\begin{quote}
\small
\textit{Once upon a time}, there was a little girl named Lily. She loved to play outside with his friends. One day, the park saw a big rock in her hand. Lily was so happy and felt very sad.

Benny was happy and proud of herself for little girl.

But then, a cool of colorful things are friends and they all wanted to help. They liked to play with his friends. One day, Tim found a big rock and the room was always happy. But then she got to school, but she always remember how to make it fit a new park in the box. They were both very sour feel sad. The doctor said that it was okay. He thought it was so funny. She would make some wood and she was so happy and proud. The end.
\end{quote}

\subsection{Task Suite Ablation Study}
\label{appendix:suite-ablation}
No-tracking-channel ablates the tracking channel $\vz$ away, leaving SRM with just the memory channel $\mH$. No-clock removes $\mK_c\vc_t$ from the key projections and $\mW_{q,c}$ from the query projections. No-previous-input-projections remove $\mK_p\vx_{t-1}$ from the key projections and $\mV_p\vx_{t-1}$ from the value projections.

With components ablated away, we give back the capacity to components that are still present, by matching FLOPs per timestep to SRM. We use the same architecture, optimizer, hyperparameter values, initialization scheme, and learning rate tuning grid as in the task-suite streaming learning experiment described in Appendix~\ref{appendix:suite-exp} and Appendix~\ref{appendix:diagnostics}. 
\label{appendix:frozen-write}

\subsection{Frozen Write Parameters Ablation Study}
\label{appendix:freeze-write-ablation}
$\mK$ denotes the concatenation of the projection weights $\mK_x$, $\mK_p$, and $\mK_c$. $\mV$ denotes the concatenation of the projection weights $\mV_x$ and $\mV_p$.

We test variants of SRM: with $\mK$ learnable, with $\mV$ learnable, with $\mK$ and $\mV$ both learnable, and with $\mK$, $\mV$ as well as their bias terms (that are functionally fixed at 0 by standard SRM) learnable. All variants use the same state size (no FLOPs matching here). We use the same architecture, optimizer, hyperparameter values, initialization scheme, and learning rate tuning grid as in the task-suite streaming learning experiment described in Appendix~\ref{appendix:suite-exp} and Appendix~\ref{appendix:diagnostics}. 

In the recall task, each sequence is 9 timesteps long, and the final query timestep shows a reserved query symbol. Symbols from the input alphabet are randomly sampled at every non-query timestep without replacement. At the final query-timestep, the learner is tasked with recalling the symbol appears at the first timestep of the sequence. A cross-entropy loss occurs at this final timestep. Figure~\ref{fig:ablation-toy} (Right) reports the evaluation accuracy at the end of training, on a held-out set of examples.

\subsection{Tracking-Channel Separation}
\label{appendix:hh-separation}
The analysis of the SRM's tracking channel in Section~\ref{subsec:ablation} uses the SRM implementation at initialization with input width $d=64$, $d_k=64$ memory rows, one read head, and value width $d_v=16$. We set $r_\text{min}=r_\text{max}=1$, so that the retention is the identity and the memory neither decays nor grows. No parameter is learned. Results are averaged over three independent runs.

Each sequence contains $4$ key--value pairs, with keys drawn without replacement and values drawn uniformly from two vocabularies of $8$ random unit vectors, followed by $\Delta$ inputs drawn uniformly from the unit sphere. The merged variant computes its axes as $\vu_t=\mW\vx_t/\lVert\mW\vx_t\rVert$ with a fixed random orthogonal $\mW\in\mathbb{R}^{d_k\times d}$, so the axes are isotropic, and uses a constant $\beta$. For the last pair, whose key $\vk_{t_0}$ is written at step $t_0$, the query is the written key $\vq=\vk_{t_0}$, which is also its position after $\Delta$ steps of SRM's identity retention. Address survival is $\vq^\top\tilde\vk/\lVert\vq\rVert^2$, where $\tilde\vk$ is $\vk_{t_0}$ propagated alone through the same transitions. Retrieval reads $\vq^\top\mH_t/\lVert\vq\rVert^2$ from the full memory and chooses the most cosine-similar of the $8$ candidate values. Each point uses $2{,}000$ sequences.

In the shell game, there are $G=5$ cups and one token per transposition, each embedded as a random unit vector in $\mathbb{R}^{64}$; the ball starts under the first cup and $\vz_0=\ve_1$. For aligned axes, we set $\mU$ so that $\mU\vx=\ve_i-\ve_j$ for the token that swaps cups $i$ and $j$, and set $\beta=2$. For random axes, $\mU$ is SRM's random semi-orthogonal projection and the strength $\beta_t=2\sigma(\vw_\beta^\top\vx_t+b_\beta)$ keeps its default initial value, which is close to $1.9$. The written variant adds $\eta\vw_t$ with $\vw_t=\mW_w\vx_t/\lVert\mW_w\vx_t\rVert$ for a fixed random $\mW_w$. The memory-only probe uses the flattened $\mH_t$ ($1{,}024$ features). Each probe is a ridge regression with bias onto the one-hot ball position, fit to standardized features of $6{,}000$ sequences split $60/20/20$ into training, validation, and test sets; the ridge coefficient is chosen from $\{10^{-4},\dots,10^{3}\}$ on the validation set, and the probe is refit on training and validation data and evaluated on the test set.

%%%%%%%%%%%%%%%%%%%%%%%%%%%%%%%%%%%%%%%%%%%%%%%%%%%%%%%
\section{SRM with RTRL}
\label{appendix:rtrl-empirical}

\paragraph{Cost Analysis. } We instantiate the architecture and its parameters identically for one-step BPTT and RTRL and measure the FLOPs of one training timestep at batch size one. We do not count the warm-up timestep that initializes recurrent sensitivities and optimizer state. A synthetic gradient at the output logits drives backward computation. The measurement counts all model layers, RTRL sensitivity maintenance, and one StreamingOptimizer update. The cost of loss computation, gradient clipping, initialization, and data movement is excluded. We count a multiply-add as two FLOPs and assign fixed formula-based costs to nonlinear operations. Table~\ref{tab:srm_training_flops} reports the resulting equivalent FLOPs per timestep for SRM training, for all models used in experiments.

\begin{table}[ht]
\scriptsize
\centering
\begin{tabular}{lrrr}
\toprule
\textbf{Model} &
\textbf{one-step BPTT} &
\textbf{RTRL} &
\textbf{Ratio} \\
\midrule
Task Suite Model & 113,796     & 201,543       & $1.771\times$ \\
MQAR small       & 629,715     & 720,729       & $1.145\times$ \\
MQAR medium      & 1,965,923   & 2,656,873     & $1.351\times$ \\
MQAR large       & 6,767,235   & 12,150,921    & $1.796\times$ \\
Language         & 259,341,436 & 1,773,230,984 & $6.837\times$ \\
\bottomrule
\end{tabular}
\caption{FLOPs per timestep for SRM training with one-step BPTT and RTRL. Costs include forward computation, backward computation, and one optimizer update.}
\label{tab:srm_training_flops}
\end{table}
Between the two streaming-learning modes, SRM with one-step BPTT matches our design principle of cheap computation requirement (D2) much better.

\begin{figure}[ht]
    \centering
    \includegraphics[width=0.9\linewidth]{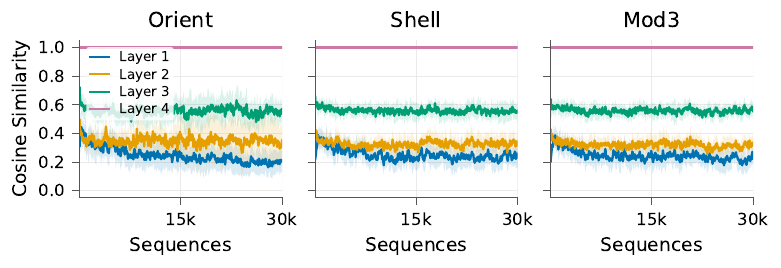}
    \caption{RTRL gradients alignment against exact full BPTT gradients throughout training.}
    \label{fig:rtrl-multilayer}
\end{figure}

\paragraph{Gradients Analysis.} In an architecture with only one recurrent layer, RTRL gradient is exact. However, in a multi-layers architecture, RTRL implementations usually track only within-layer forward-mode sensitivities and neglect cross-layer sensitivity terms. Consequently, the computation of the gradients that flow through layers is an approximation. We quantify this discrepancy by measuring, at each layer, the cosine similarity between the gradients computed by RTRL and the exact gradients computed by full BPTT. All computations are performed in double precision. Figure~\ref{fig:rtrl-multilayer} shows the result for an architecture of four SRM layers, each preceded by RMSNorm with a detached normalization scale. Detaching the normalization scale preserves its forward stabilizing effect while excluding gradients through the scale from the gradient alignment comparison.

For the top layer (Layer 4), the gradients estimated by RTRL match those computed by full BPTT throughout training. However, the approximation deviation received by the next layer is already significant. The gradients that the bottommost layer (Layer 1) receives, as estimated by RTRL, only has a cosine similarity of around 0.2, to the gradients computed by full BPTT.

\end{document}

%% file: math_commands.tex
\usepackage{amsmath,amsfonts,bm}

\def\eqref#1{equation~\ref{#1}}
\def\1{\bm{1}}

\def\va{{\bm{a}}}
\def\vb{{\bm{b}}}
\def\vc{{\bm{c}}}

\def\ve{{\bm{e}}}

\def\vg{{\bm{g}}}
\def\vh{{\bm{h}}}

\def\vk{{\bm{k}}}

\def\vo{{\bm{o}}}

\def\vq{{\bm{q}}}
\def\vr{{\bm{r}}}

\def\vu{{\bm{u}}}
\def\vv{{\bm{v}}}
\def\vw{{\bm{w}}}
\def\vx{{\bm{x}}}
\def\vy{{\bm{y}}}
\def\vz{{\bm{z}}}

\def\mA{{\bm{A}}}
\def\mB{{\bm{B}}}

\def\mH{{\bm{H}}}
\def\mI{{\bm{I}}}
\def\mJ{{\bm{J}}}
\def\mK{{\bm{K}}}

\def\mR{{\bm{R}}}
\def\mS{{\bm{S}}}

\def\mU{{\bm{U}}}
\def\mV{{\bm{V}}}
\def\mW{{\bm{W}}}

\DeclareMathAlphabet{\mathsfit}{\encodingdefault}{\sfdefault}{m}{sl}
\SetMathAlphabet{\mathsfit}{bold}{\encodingdefault}{\sfdefault}{bx}{n}

\newcommand{\dd}{\mathrm{d}}
\DeclareMathOperator{\diag}{diag}
\DeclareMathOperator{\Diag}{Diag}

%% file: figures/streaming-learner-tikz.tex
% colours live in the body: \includestandalone discards this file's preamble
\colorlet{ink}{black!80}
\colorlet{boxfill}{black!11}
\colorlet{boxdraw}{black!45}
\colorlet{panelfill}{black!6}
\colorlet{paneldraw}{black!50}
\colorlet{panellabel}{black!55}
\begin{tikzpicture}[
    font=\sffamily\small, line cap=round,
    >={Latex[length=1.7mm,width=1.3mm]},
    wire/.style={->, line width=0.6pt, ink},
    upd/.style={->, line width=0.6pt, draw=gray, opacity=0.45,
                dash pattern=on 2.2pt off 1.6pt},
    op/.style={rectangle, rounded corners=5pt, draw=boxdraw, fill=boxfill, line width=0.7pt,
               minimum size=0.95cm, inner sep=0pt},
    note/.style={font=\sffamily\footnotesize, black},
  ]

  % model
  \node[op] (f) at (0,0) {};
  \node[op] (g) at (2.5,0) {};
  \draw[wire] (f) -- node[pos=0.3, above, font=\footnotesize] {$\bm{h}_t$} (g);

  % learning algorithm
  \node[draw=boxdraw, fill=boxfill, rounded corners=6pt, line width=0.7pt,
        minimum width=3.5cm, minimum height=0.68cm] (alg) at (1.25,1.3) {learning algorithm};
  \coordinate (state-split) at ($(f)!0.4!(g)$);
  \fill[ink] (state-split) circle (1.1pt);
  %\draw[wire] (state-split) -- (alg.south);
  \coordinate (state-return) at (-0.92,0.12);
  %\coordinate (state-input) at ($(f.west)+(0,0.12)$);
  %draw[wire, rounded corners=2pt] ($(alg.west)+(0,-0.12)$) -- (-1.45,1.18) -- (state-return) -- (state-input);
  \draw[wire, rounded corners=5pt] (state-split)
%|- ($(f.south)+(0,-0.47)$) -- (f.south);
|- ($(f.south)+(-1,-0.47)$)
|- ($(f.west)+(0,-0.12)$);
\node[draw=boxdraw, fill=boxfill, rounded corners=2pt, inner sep=1pt, font=\fontsize{4}{5}\selectfont] at ($(f.south)+(0.,-0.47)$) {delay};
  \node[above=2pt, inner sep=0pt, font=\footnotesize]
%    at ($(state-return)!0.5!(state-input)$) {$\bm{h}_{t-1}$};
at ($(f.south)+(-0.6,-0.32)$) {$\bm{h}_{t-1}$};
  \draw[upd] ($(alg.south -| f)+(-0.25,0)$)
    -- ($(f.north)+(-0.25,0.12)$) -- ($(f.south)+(0.35,-0.12)$);
  \draw[upd] ($(alg.south -| g)+(-0.25,0)$)
    -- ($(g.north)+(-0.25,0.12)$) -- ($(g.south)+(0.35,-0.12)$);
  \node[inner sep=0pt] at (f.center) {$f_{\bm{\theta}}$};
  \node[inner sep=0pt] at (g.center) {$g_{\bm{\phi}}$};

  % learner boundary
  \coordinate (loopbot) at (0,-0.9);
  \coordinate (rightpad) at ($(g.east)+(0.35,0)$);
  \begin{scope}[on background layer]
    \node[draw=paneldraw, fill=panelfill, rounded corners=10pt, line width=0.6pt,
          inner xsep=14pt, inner ysep=6pt, fit=(f)(g)(alg)(loopbot)(state-return)(rightpad)] (learner) {};
  \end{scope}
  \node[anchor=south west, font=\sffamily\footnotesize, panellabel] at ($(learner.south east)+(-35pt,0pt)$) {Learner};

  % incoming streams
  \foreach \name/\y/\sym in {x/0.12/\bm{x}_t, y/1.42/\bm{y}_t} {
    \node[inner xsep=1pt] (\name0) at ($(learner.west |- 0,\y) + (-1.0,0)$) {$\sym$};
  }
  \node[note, inner xsep=0pt, left=3pt of x0] {input};
  \node[note, inner xsep=0pt, left=3pt of y0] {feedback};
%  \draw[wire] (x0.east) -- ($(f.west)+(0,-0.12)$);
%  \draw[wire, rounded corners=3pt] (-1.0,-0.12) |- ($(alg.west)+(0,-0.12)$);
%  \fill[ink] (-1.0,-0.12) circle (1.1pt);
\draw[wire] (x0.east) -- ($(f.west)+(0,0.12)$);
\draw[wire, rounded corners=3pt] (-1.0,0.12) |- ($(alg.west)+(0,-0.12)$);
\fill[ink] (-1.0,0.12) circle (1.1pt);
  \draw[wire] (y0.east) -- ($(alg.west)+(0,0.12)$);

  % output, also seen by the learning algorithm
  %\coordinate (split) at ($(learner.east |- g) + (0.45,0)$);
  %\draw[wire] (g.east) -- ($(split)+(0.75,0)$);
  %\node[note, inner xsep=0pt, anchor=west] (output-label) at ($(split)+(0.85,0)$) {output};
\coordinate (split) at ($(g.east)+(0.45,0)$);
\draw[wire] (g.east) -- ($(learner.east |- g)+(1.20,0)$);
\node[note, inner xsep=0pt, anchor=west] (output-label) at ($(learner.east |- g)+(1.30,0)$) {output};
  \node[inner xsep=1pt, right=3pt of output-label] {$\bm{o}_t$};
  %\fill[ink] (split) circle (1.1pt);
  %\draw[wire, rounded corners=2pt] (split) |- (alg.east);
\end{tikzpicture}